\documentclass[11pt]{article}

\usepackage[final]{acl}

\usepackage[utf8]{inputenc}
\usepackage{tcolorbox}
\usepackage{pmboxdraw} 
\usepackage{amssymb}   

\usepackage{times}
\usepackage{latexsym}
\usepackage{enumitem}
\usepackage{booktabs}
\usepackage{multirow}
\usepackage{graphicx}
\usepackage{amsmath}
\usepackage{algorithm}
\usepackage{algorithmic}
\usepackage[T1]{fontenc}

\usepackage[utf8]{inputenc}
\usepackage[normalem]{ulem}
\usepackage{microtype}

\usepackage{inconsolata}

\usepackage{tabularx}
\usepackage{graphicx}

\newcommand{\ours}{\textsc{Metro}}
\title{\ours: Towards Strategy Induction from Expert Dialogue Transcripts \\for Non-collaborative Dialogues}

\author{Haofu Yang$^{4-6}$,
Jiaji Liu$^{2}$, 
Chen Huang$^{1\dag}$,
\textbf{Faguo Wu}$^{4-6\dag}$,
\textbf{Wenqiang Lei}$^{3,7}$,
\textbf{See-Kiong Ng}$^{1}$ \\
$^{1}$ National University of Singapore \quad$^{2}$ Nankai University \quad
$^{3}$ Sichuan University  \\
$^{4}$ School of Artificial Intelligence, Beihang University \\ 
$^{5}$ Beijing Advanced Innovation Center for Future Blockchain and Privacy Computing \\
$^{6}$ Key Laboratory of Mathematics, Informatics and Behavioral Semantics (LMIB) \\
$^{7}$ Engineering Research Center of Machine Learning and Industry Intelligence, \\Ministry of Education, China\\
  }

\begin{document}
\maketitle

\def\thefootnote{$\dag$}\footnotetext{Corresponding authors: huang\_chen@nus.edu.sg\par
\hspace*{0.2cm} faguo@buaa.edu.cn}\def\thefootnote{\arabic{footnote}}

\begin{abstract}
Developing non-collaborative dialogue agents traditionally requires the manual, unscalable codification of expert strategies. We propose \ours, a method that leverages large language models to autonomously induce both strategy actions and planning logic directly from raw transcripts. \ours\ formalizes expert knowledge into a Strategy Forest, a hierarchical structure that captures both short-term responses (nodes) and long-term strategic foresight (branches). Experimental results across two benchmarks show that \ours\ demonstrates promising performance, outperforming existing methods by an average of 9\%-10\%. Our further analysis not only reveals the success behind \ours\ (strategic behavioral diversity and foresight), but also demonstrates its robust cross-task transferability. This offers new insights into building non-collaborative agents in a cost-effective and scalable way. Our code is available at \url{https://github.com/Humphrey-0125/METRO}.
\end{abstract}

\section{Introduction}
Building conversational agents for non-collaborative dialogues is receiving increased attention \cite{zhang2024strength, hu2025astro, deng2023plug, deng-rethinking}, particularly in applications like price negotiation \cite{he2018decoupling}, donation persuasion \cite{wang-etal-2019-persuasion}, or debt collection \cite{zhang-etal-2023-towards-effective}. In these interactions, the user (i.e., the opposing party) holds conflicting interests with the agent and consequently exhibits resistance during the multi-turn conversation \cite{deng-etal-2023-goal, lei2022interacting, shi2019build, dutt2021resper}. 
This dynamic requires the agent to employ effective dialogue strategies, securing an agreement favorable to itself \cite{zhan2024let}.

Effective dialogue strategies are typically latent within expert dialogue transcripts of non-collaborative scenarios. These transcripts implicitly encode both \underline{strategy actions} (what to say, e.g., \textit{emotional appeals} or \textit{foot-in-the-door}) and \underline{planning logic} (the timing of these actions, e.g., \textit{when to employ an emotional appeal}). Consequently, extracting these strategies has always been a labor- and knowledge-intensive task \citep{N4-append,fransen2015strategies}. In this case, to build a non-collaborative dialogue agent, the standard paradigm relies on domain experts to first manually analyze transcripts and codify a set of strategy actions \cite{he2018decoupling, wang-etal-2019-persuasion,bonaldi-etal-2022-human,zhang-etal-2023-towards-effective}, followed by the strategy planner training using pre-collected or LLM-roleplayed transcripts \cite{zhang2024strength, deng2023plug, yu-etal-2023-prompt}. However, this process is inherently expensive and unscalable. With the emergence of Large Language Models (LLMs) capable of powerful induction\footnote{Synthesize broader patterns from granular examples.} \cite{cai2025role, de2025inductive}, there is a compelling opportunity to shift such paradigm: \textbf{Can we leverage LLMs to induce both strategy actions and planning logic directly from raw transcripts?}

To this end, we propose \textbf{\ours}\ (\uline{M}ulti-dim\uline{E}ntional s\uline{TR}ategy induction from dial\uline{O}gue transcripts), a novel method that extracts and formalizes raw dialogue transcripts into a "Strategy Forest". Specifically, \ours~operates in two phases: Induction and Application. During the Induction phase, \ours\ extracts strategy actions as nodes and organizes multi-turn action trajectories into trees, each conditioned on specific dialogue states. This hierarchical architecture captures planning logic across two temporal scales: immediate child nodes represent short-term tactical responses, while full branches encapsulate long-term strategic foresight. By synthesizing these perspectives, \ours\ creates a multi-dimensional representation of the strategies embedded within expert transcripts. In the subsequent Application phase, \ours\ analyzes the current dialogue state and queries the Strategy Forest to retrieve the most relevant tree, which serves as a blueprint to guide the agent’s decision-making. As such, \ours\ enables cost-effective, scalable, and strategic performance by effectively leveraging historical expert knowledge.


We conduct experiments on two benchmark datasets to demonstrate the superiority of \ours. Specifically, \ours~achieves significant performance gains, surpassing other induction methods by an average of 10.24\% across both benchmarks. Our analysis attributes this success to the generation of diverse strategic actions and informative planning logic, which endow agents with a richer behavioral patterns and strategic foresight. Moreover, \ours~exhibits robust scalability via cross-task generalization: strategies induced from external domains remain highly effective in the target setting. 
Finally, we validate the potential of LLM-synthesized transcripts as a viable alternative to expert transcripts. To sum up, we conclude our main contributions as follows.
\begin{itemize}[leftmargin=*]
\item We highlight strategy induction from expert transcripts as an effective and scalable solution for non-collaborative dialogues.
\item We pioneer \ours, which formalizes expert knowledge in transcripts into a Strategy Forest, representing strategy actions and planning logic.
\item We conduct extensive experiments to validate the effectiveness of \ours\ and provide an in-depth analysis of the diversity, transferability, and expertise-sensitivity of the induced strategies.
\end{itemize}

\section{Related Work}

\noindent\textbf{LLM Agents for Non-collaborative Dialogues}. Effective dialogue strategies are the key of non-collaborative agents. Traditionally, these agents rely on expert-defined action sets derived from manual analysis of dialogue transcripts, such as action sets for persuasion \cite{wang-etal-2019-persuasion}, counterspeech \cite{bonaldi2022human}, and negotiation \cite{he2018decoupling}. While these actions remain widely used \cite{wang-etal-2025-battling,zhang2024strength, deng2023survey, yu-etal-2023-prompt}, they represent a labor-intensive bottleneck. Furthermore, deploying these actions requires sophisticated planning. Because simple prompting often lacks fine-grained control \cite{deng2023plug}, recent works have introduced external planners like MCTS \cite{yu-etal-2023-prompt} or specialized plug-in models \cite{deng2023plug, zhang2024strength, wang-etal-2025-battling, du-etal-2025-planning}. Despite these advances, the overall pipeline remains resource-intensive and unscalable. While PRINCIPLES \cite{kim2025principles} automates the extraction of planning logic (\textit{when [situation], you should [action A], rather than [action B], because [reason]}), it fails to capture multi-turn planning logic essential for non-collaborative setting.

\noindent\textbf{Expert Knowledge Induction from Dialogues}. Inductive learning empowers LLMs to synthesize broader patterns from granular examples \cite{cai2025role, de2025inductive,huang2024araida,louie-etal-2024-roleplay}. This capability has catalyzed significant research into task-specific dialogue transcripts, which act as rich repositories of human experience and knowledge. In particular, researchers extract emergent user intents and slots \cite{liang-etal-2024-survey, liang2024synergizing, vukovic2024dialogue, finch2025generative}, and derive dialogue structures \cite{burdisso-etal-2024-dialog2flow, raedt2024revisiting, choubey2025turning, pryor2023using}. Furthermore, recent studies focus on identifying strategy actions \cite{ou2024inductive, wang2025dystil,he2025simulation, liu2025exploring, hu2025astro}, together with the rationales of these actions governing the planning logic \cite{jiang2025chatmap, xie2024few}. 
While \citet{kim2025principles} recently adapts these ideas to non-collaborative dialogues, existing methods treat strategies as independent units. Instead, we induce and structure dialogue strategies into a forest, capturing the complex dynamics. 



\begin{figure*}
    \centering
    \includegraphics[width=0.99\textwidth]{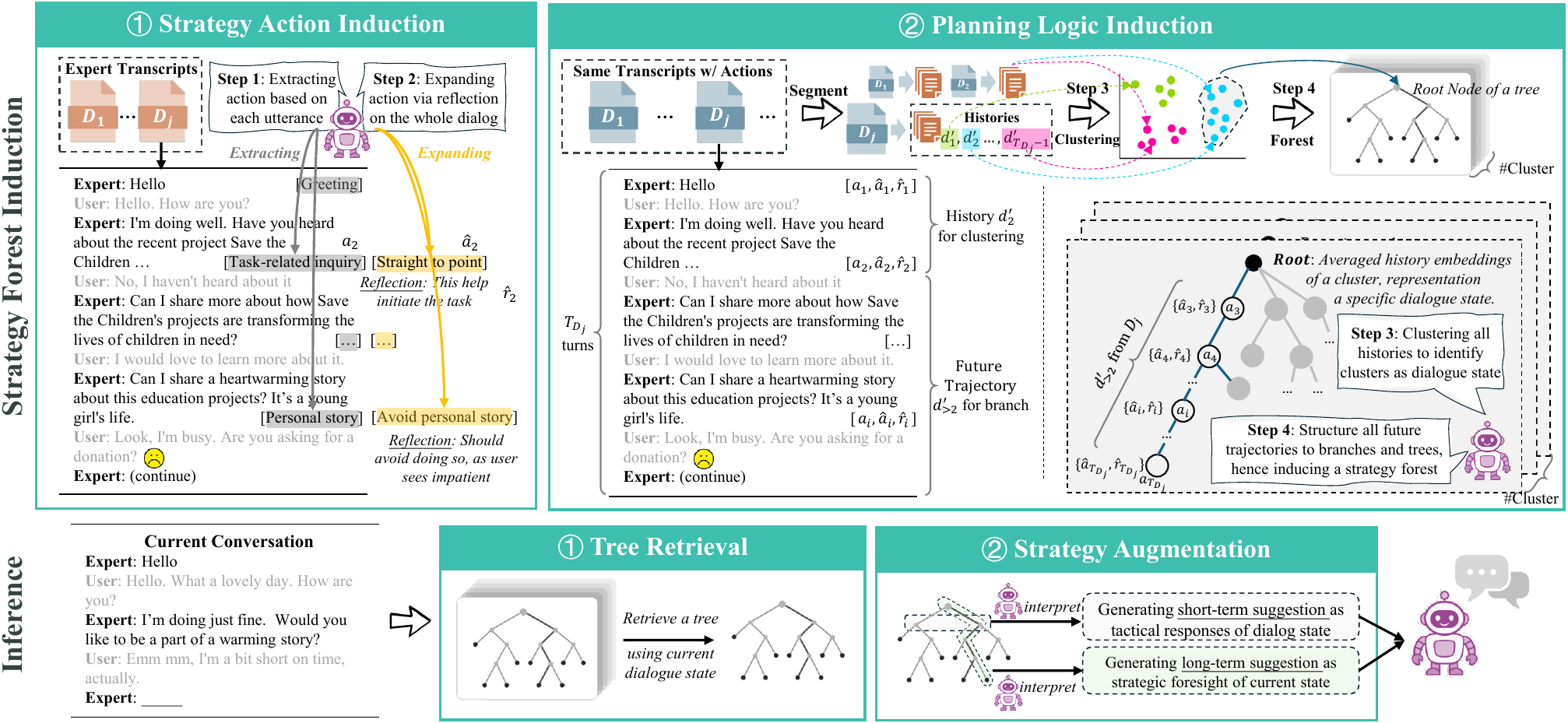}
    \caption{Overview of \ours. It formalizes expert knowledge within transcripts into a Strategy Forest, representing strategy actions and planning logic in a unified structure.}
    \label{fig:main}
\end{figure*}

\section{\ours}
As illustrated in Figure \ref{fig:main}, \ours~first conducts an offline induction of the Strategy Forest directly from raw dialogue transcripts (cf. Section \ref{induction}), subsequently utilizing this structure to guide inference within non-collaborative dialogues (Section \ref{inference}). 

\noindent\textbf{Notations \& Overview}. Existing methods rely on a fixed set of human-crafted strategic actions $\mathcal{A}$. Let $d_t = (v_1, v_1^{\text{usr}}, \dots, v_t, v_t^{\text{usr}})$ denote the dialogue history at turn $t$, where $v_i^{\text{usr}}$ is the $i$-th user utterance and $v_i$ is the $i$-th agent utterance corresponding to a chosen action $a_i \in \mathcal{A}$. Typically, a trained planner takes $d_t$ as input to predict the next action $a_{t+1} \in \mathcal{A}$. In contrast, \ours~directly induces dialogue strategies from a corpus of historical transcripts $\mathcal{D}$. Each transcript $D \in \mathcal{D}$ is defined as a sequence $D = (u_1, u_1^{\text{usr}}, \dots, u_{T_D}, u_{T_D}^{\text{usr}})$, where $u_i$ is $i$-th utterance from expert and $T_D$ represents the total number of turns in transcript $D$. \ours~transforms $\mathcal{D}$ into a Strategy Forest $\mathcal{F}$ and induces an expanded action set $\mathcal{A}'$ (where $\mathcal{A} \subseteq \mathcal{A}'$). Within this forest, the root node of the $i$-th tree $f_i \in \mathcal{F}$ represents a specific dialogue state $S_i$, while all subsequent nodes encode effective strategic actions $a' \in \mathcal{A}'$ from that state. Structurally, the $j$-th branch extending from the root to a leaf node represents an action trajectory $\text{traj}_{ij} \in f_i$, which encodes long-term strategic value. Conversely, the immediate children of the root node, denoted as $\text{childro}_i$, capture short-term value, indicating the effective next step to take from the root state $S_i$. During inference, given the Strategy Forest $\mathcal{F}$ and the current dialogue history $d_t$, \ours~first retrieves the most relevant tree $f \in \mathcal{F}$, identifying the root node whose dialogue state $S$ bears the highest semantic similarity to $d_t$. Subsequently, \ours~aggregates the top-$K$ $\text{childro}_i$ and $\text{traj}_{ij}$ within $f$ to inform a next-step decision.


\subsection{Strategy Forest Induction}
\label{induction}

\subsubsection{Strategy Action Induction}
Standard knowledge induction relies on simply extracting target information from raw transcripts, such as query intents \cite{gung-etal-2023-intent, liang2024synergizing} and dialogue flows \cite{burdisso-etal-2024-dialog2flow, raedt2024revisiting, choubey2025turning}. However, this method is limited by the quality and diversity of the raw transcripts. To address this, we propose augmenting the extraction process with the internal knowledge of LLM agents. Specifically, by using strategy actions derived from raw transcripts as bases, we prompt the LLM agent to generate refined new strategic actions, effectively transcending the limitations of the original transcripts.

\noindent\textbf{Action Extraction}. \ours~starts with the identification of the strategic action $a_i$ for every expert utterance $u_i$ in transcript $D$. To streamline this, \ours~supports two modes of operation. If an expert-curated action taxonomy exists for the target non-collaborative task\footnote{There are many human-crafted action taxonomies \cite{wang-etal-2019-persuasion, bonaldi2022human, he2018decoupling}}, \ours~simply assigns the most appropriate action from the existing set. In the absence of such a taxonomy, \ours~leverages its LLM backbone to autonomously induce a domain-specific action set based on the task description, following \cite{hu2025astro}.
\begin{equation}
    a_i = \text{Extraction}(u_i),  \forall u_i \in D, D\in \mathcal{D}.
\end{equation}

\noindent\textbf{Action Expansion}. Using its LLM backbone, \ours~expands the strategy actions via self-reflection. For each turn $i$ of transcript $D$, denote the dialogue history $d'_i = (u_1, u_1^{\text{usr}}, \dots, u_i, u_i^{\text{usr}})$. Taking the strategic action $a_i$ and history $d'_i$ as input, \ours~assesses the quality of the expert utterance $u_i$ by analyzing the user's reaction $u_i^{\text{usr}}$ (determined by LLM-as-judge). For effective utterance $u_i$, \ours~summarizes an new action $\hat{a}_i^0$ without access $a_i$ to enhance generalization (e.g., \textit{elicit empathy to deepen engagement}). For ineffective one, the original action $a_i$ is flagged as a `\textit{should-not-do}' action $\hat{a}_i^1$ for the given context, such as `\textit{When [situation], you must avoid [$a_i$]}'.
This guides the agent away from poor choices.  
\begin{equation}
    (\hat{a}_i, \hat{r}_i) = \text{Expansion}(d'_i, a_i).
\end{equation}
Finally, each $u_i$ is augmented with two actions: the original $a_i$ and the LLM-expanded $\hat{a}_i \in \{\hat{a}_i^0, \hat{a}_i^1 \}$, accompanied by a rationale $\hat{r}_i$ for $\hat{a}_i$.

\subsubsection{Planning Logic Induction}
A full action trajectory is encoded within a dialogue transcript. It is defined as the sequence of expert strategy actions, inferred from the expert's utterances and labeled with the final task outcome (success or failure). \ours~translates these trajectories using a state-centric tree structure, where trees with different dialogue states form the forest. To achieve this, we decompose each trajectory into local sub-trajectories, where each sub-trajectory represents an action path initiating from a specific dialogue state and carrying a final task outcome. Finally, we aggregate all sub-trajectories rooted in the identical dialogue state (regardless of their source transcript) into a unified tree. This synthesizes diverse historical planning logic within transcripts to enhance generalization.

\noindent\textbf{Dialogue State Identification \& Clustering}. We process each transcript $D$ (representing a full trajectory of length $T_D$) by segmenting it into several past histories $\{d'_1, ..., d'_{T_D-1}\}$ and their corresponding future trajectories $\{d'_{>1}, ..., d'_{>T_D-1}\}$, where $d'_{i}$ denotes the dialogue history up to turn $i$ (inclusive) and $d'_{>i}$ represents all subsequent turns. We posit that these textual histories $d'_{i}$ provide a denser semantic basis for state modeling than strategic actions alone. Therefore, we convert each history $d'_i$ into a semantic embedding\footnote{We use \textit{bge-large-en-v1.5} \cite{bge_embedding}.} $\text{Emb}(d'_i)$ to represent the dialogue state preceding utterance $u_{i+1}$. We then apply K-Means to cluster all embeddings sourced from different transcripts. 
As a result, each expert utterance $u_i$ ($i\neq 1$) within transcript $D$ is associated with its preceding state embedding $\text{Emb}(d'_{i-1})$, a cluster index, the future trajectory $d'_{>i-1}$, and its action metadata $\{a_i, \hat{a}_i, \hat{r}_i\}$.


\noindent\textbf{Strategy Forest Induction}. \ours~induces a Strategy Forest by constructing a state-centric tree from each cluster. In this forest, the root node of each tree represents a distinct dialogue state, and its branches encode effective planning logic from that state. To achieve this, given a cluster, we first define the cluster centroid as the root node, representing the dialogue state in the embedding space. The branches are then formed using all future trajectories within that cluster. Since not all branches are equally valuable, we prune low-value branches and retain only the top-K branches with the highest value. This process is detailed below.
\begin{itemize}[leftmargin=*]
    \item \uline{Branch Generation}. We construct branches for a tree by organizing all future trajectories from the same cluster. Instead of using $\hat{a}_i$, nodes are represented by the original data-driven actions $a_i$ to ensure authenticity, as they form trajectories that are grounded in real user interactions. Identical action sequences are merged to form consolidated branches (e.g., \textit{Donation information $\to$ Foot in the door $\to$ Emotion appeal}).
    
    \item \uline{Node Value Estimation}. We assign a value to every node along a branch. This value is primarily determined by the final task outcome\footnote{Such as success rate and Sale-to-List Ratio (SL\%).} of the dialogues that passed through it, a process analogous to value backpropagation in MCTS \cite{silver2016mastering} (cf. Appendix \ref{sec:planning_logic_induction_impl}).

    \item \uline{Branch Value Estimation \& Pruning}. Besides branches leading to failures, a branch that appears successful but has been observed only a few times may also be unreliable. To mitigate the risk of overestimating sparsely observed but successful trajectories, our branch valuation incorporates both performance and statistical confidence. Each node is augmented by a confidence term. This is derived from the Wilson Score Lower Bound \cite{wilson1927probable} using the empirical success and visitation counts of each node (cf. Appendix \ref{sec:planning_logic_induction_impl}). The final value of a branch is then defined as the weighted sum of its constituent node values and their corresponding confidence scores. Subsequently, we perform pruning by retaining the top-K branches with the highest value. This is further optimized in practice using Beam Search, accelerating the pruning process. 
    

\end{itemize}

\begin{table*}[]
\centering
\resizebox{0.99\textwidth}{!}{%
\begin{tabular}{l|llllllllll}
\toprule
\multirow{2}{*}{\textbf{Methods}} & \multicolumn{2}{c}{\textbf{P4G}} & \multicolumn{3}{c}{\textbf{CB}} & \multicolumn{2}{c}{\textbf{P4G*}} & \multicolumn{3}{c}{\textbf{CB*}} \\ \cmidrule(lr){2-3} \cmidrule(lr){4-6} \cmidrule(lr){7-8} \cmidrule(lr){9-11} 
 & \textbf{SR} $\uparrow$ & \textbf{AT} $\downarrow$& \textbf{SR} $\uparrow$ & \textbf{AT} $\downarrow$ & \textbf{SL\%} $\uparrow$ & \textbf{SR} $\uparrow$ & \textbf{AT} $\downarrow$& \textbf{SR} $\uparrow$ & \textbf{AT} $\downarrow$& \textbf{SL\%} $\uparrow$ \\ \midrule
Standard \cite{deng2023prompting}& 0.620 & 4.56 &0.185 & 5.89&0.154 &0.333 &5.20 &0.283 &5.65 &0.285 \\
Proactive \cite{deng2023prompting}&0.660 & \uline{4.11} &0.170 & 6.03 &0.116 &0.400 &\textbf{4.67} &0.283 &5.41 &0.198 \\
ProCoT \cite{deng2023prompting}&0.680 & 4.29&0.315 &6.46 &0.186 &0.533 &5.13 &0.333 &6.10 &0.282 \\
ICL-AIF \cite{fu2023improving}& 0.615& \textbf{3.89}& 0.325&6.11 &\textbf{0.231} &0.583 &5.09 &0.450 &6.00 &0.348 \\ \midrule
GDP-Zero \cite{yu-etal-2023-prompt}&0.660 &5.35  &\uline{0.495} & 6.16& 0.125&0.600 & 5.11&0.450 &6.37 &0.377 \\
PPDPP \cite{deng2023plug}&0.730 &4.67 &0.250 &\uline{5.62} &0.150 &\uline{0.633} &6.13 &0.383 &\textbf{4.30} &0.215 \\
\midrule
PRINCIPLES \cite{kim-etal-2025-principles}&\uline{0.770} & 5.24& 0.485& 5.73& 0.149&0.600 &\uline{4.78} &\uline{0.467} &5.86 &\uline{0.389} \\
\ours (\textit{ours}) &\textbf{0.780} & 4.76&\textbf{0.575} &\textbf{5.38} & \uline{0.189} &\textbf{0.661}&5.10 & \textbf{0.483}&\uline{5.21} &\textbf{0.411} \\ \bottomrule
\end{tabular}%
}
\caption{Evaluation results for baselines and \ours~across two datasets, incorporating both user simulation (P4G \& CB) and human interaction studies (P4G* \& CB*). Appendix \ref{moreex} details performance on diverse users.}
\label{tab:main}
\end{table*}

Ultimately, each tree inherently captures planning logic at two distinct temporal resolutions: 
\begin{itemize}[leftmargin=*]
    \item \uline{Breadth Planning Logic}. Immediate child nodes, containing both data-driven and LLM-expanded actions, of the root represent short-term tactical responses conditioned on the specific state.
    \item \uline{Depth Planning Logic}. Complete branches extending from the root encapsulate effective long-term strategic foresight for that state.
\end{itemize}

\subsection{Inference with Strategy Forest}
\label{inference}
Since the induced strategy forest contains rich strategy actions and planning logic, \ours~omit the model training and directly performs retrieval augmentation during the inference to enhance its ability in non-collaborative dialogues.

\noindent\textbf{Tree Retrieval}. Before generating the response for turn $t$, \ours~encodes the current dialogue history $d_{t-1}$ into a semantic embedding $\text{Emb}(d_{t-1})$. This embedding serves as a query to traverse the Strategy Forest: we compute the cosine similarity between $\text{Emb}(d_{t-1})$ and the root node of every tree, retrieving the single tree that exhibits the highest semantic similarity to the current state.

\noindent\textbf{Strategy Augmentation}. We prompt \ours~by integrating both breadth and depth planning logic derived from the retrieved tree. Inspired by \citet{kim2025principles}, we recognize that retrieved logic may not perfectly align with the nuances of the current dialogue. Therefore, we task the LLM backbone with generating contextual strategy as suggestion.

\begin{itemize}[leftmargin=*, itemindent=0.05cm, itemsep=-2pt]
    \item \uline{Short-Term Suggestion}. the LLM reinterprets the actions from breadth logic (e.g., `\textit{provide specific and concrete information}') to fit the current dialogue context, summarizing them into a concise next-step strategy (e.g., `\textit{explaining the charity’s concrete programs and impact to address the persuadee’s curiosity}'). For simplicity, we exclusively utilize the LLM-expanded actions for this step, as they represent optimized abstractions of the original data-driven actions. 
    \item \uline{Long-Term Suggestion}. We identify the single branch with the highest average node value to serve as the depth logic (e.g., `\textit{build credibility $\to$ propose a donation}'). Similarly, the LLM reinterprets this trajectory, summarizing it into a high-level planning directive that emphasizes cumulative temporal effects (e.g., `\textit{progressively strengthening trust and commitment prior to making a donation request}'). 
\end{itemize}
Finally, these two suggestions are concatenated to guide \ours~in generating the final response. Notably, \ours\ diverges from MCTS, as it leverages multiple trees to capture multi-dimensional strategies (breadth and depth) without requiring expensive test-time computation. 

\section{Experiments}
\subsection{Experimental Setup}
\label{user_simulators}

Following common practice in the field \cite{zhang2024strength, deng2023plug, hu2025astro}, we evaluate agents by pairing them with synthetic user simulators exhibiting diverse non-collaborative behaviors. In each interaction, both the agent and simulator alternate turns, employing strategies to maximize their respective self-interests. The dialogue proceeds until the task goal is met or the maximum number of turns (i.e.,8) is reached, after which we assess the agent's performance based on the collected interactions. Finally, implementation and evaluation details are presented in Appendix \ref{imp} and \ref{evl}, respectively. Additional experimental analyses is in Appendix \ref{moreex}.

\begin{figure}[t]
    \centering
    \includegraphics[width=0.42\textwidth]{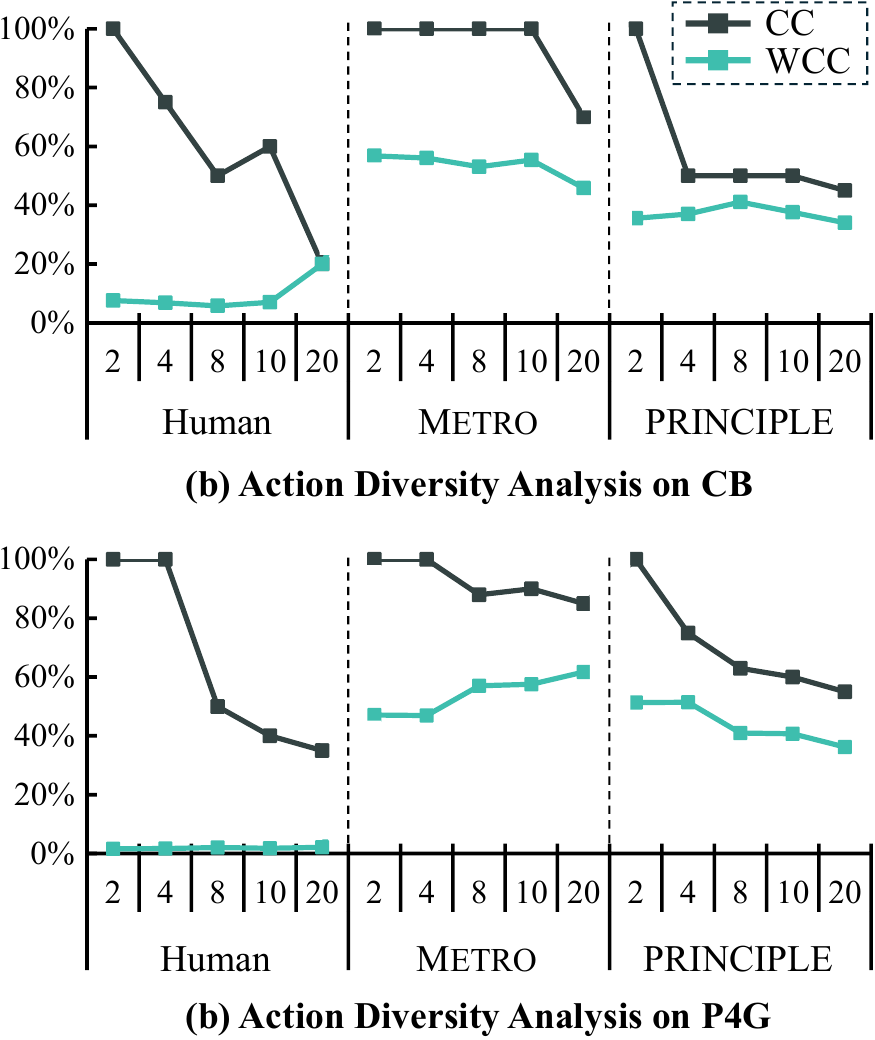}
    \caption{Action diversity analysis.}
    \label{fig:actiond}
\end{figure}

\noindent \textbf{Evaluation Tasks}. In line with \citet{deng2023prompting, wang-etal-2019-persuasion,hu2025astro,zhang2024strength}, we employ two benchmark tasks: the price negotiation task, utilizing the test dataset of CraigslistBargain (CB) \cite{he2018decoupling}, the charity persuasion task, employing the test dataset of PersuasionForGood (P4G) \cite{wang-etal-2019-persuasion}. In these scenarios, the conversational agents assume the roles of buyer and persuader, respectively, aiming to achieve their task-specific goals. 

\noindent \textbf{Baselines}.
We compare with two types of baselines. 
1) \underline{Prompt-based}, including \textit{Standard} \cite{deng2023prompting}, \textit{Proactive} \cite{deng2023prompting}, \textit{ProCoT} \cite{deng2023prompting} and \textit{ICL-AIF} \cite{fu2023improving}, which use vanilla prompts, mixed-initiative prompts, CoT, and AI feedback to select next strategies, respectively.
2) \underline{External strategy planners}, including \textit{GDP-Zero} \cite{yu-etal-2023-prompt} and \textit{PPDPP} \cite{deng2023plug}, which utilize a MCTS and a trainable plug-in for determining next-step strategies, respectively. 
3) \underline{Strategy induction}, PRINCIPLES \cite{kim-etal-2025-principles}, which utilize past conversation histories to extract dialogues strategies.


\noindent \textbf{User Simulators \& Human Participants}. Consistent with prior research \cite{deng2023survey, bianchi2024well, huang2024conceptevaluationprotocol, zhang2024strength}, we utilize diverse LLM-based user simulators, which demonstrate non-collaborative behaviors driven by specific personas. 
Following \cite{wang-etal-2019-persuasion, jiang2024evaluating, zhang2024strength}, we incorporate two distinct persona dimensions: the \textit{Big-Five Personality} \cite{goldberg1992development} and the \textit{Decision-Making Styles} \cite{scott1995decision}, augmenting each combination with an LLM-generated description. This yields a total of 200 diverse simulators per task, distributed across the 20 distinct categories (i.e., Big-Five $\times$ Styles). Furthermore, we employ 5 humans with varying personas and conduct a human study in which participants interact with the agents to assess their practical effectiveness, following existing setup \cite{deng2023prompting,zhang-etal-2024-clamber,chen-etal-2024-style}. See Appendix \ref{app:human_evaluation} for human study details.

\noindent\textbf{Strategy Forest Induction}. We utilize two expert human-human dialogue transcripts. Specifically, for CB task, we utilize the \textsc{CraigslistNegotiation} dataset, containing 6,682 dialogues \cite{he2018decoupling}. As for P4G task, we utilize \textsc{PersuasionForGood} dataset, with 1,017 dialogues \cite{wang-etal-2019-persuasion}. Using these transcripts as input, \ours~extracts strategy forest for subsequent experimentation. Unless otherwise specified, our evaluation employs strategies induced from the task-specific dataset. Crucially, as our evaluation involves real-time online interactions with users, there is no risk of data leakage from the static induction transcripts.

\noindent \textbf{Evaluation Metrics}. 
\label{preliminary_metrics}
Following \citet{deng2023plug, hu2025astro, zhang2024strength}, we consider three commonly used automated metrics: \underline{Success Rate} (SR), \underline{Average Turn} (AT) and \underline{Sale-to-List Ratio} (SL\%). Notably, SL\% \cite{zhou2019augmenting} is design specifically for the CB task to determine the effectiveness of goal completion. See Appendix \ref{detailmetri} for details.

\subsection{Main Results}
Table \ref{tab:main} validates effectiveness of \ours\ and the empirical quality of its induced strategies. 

\noindent\textbf{\ours~achieves promising performance and outperforms other induction methods across diverse benchmarks and metrics}. As shown in Table \ref{tab:main} (left), \ours\ demonstrates promising performance on the SR metric across all evaluated non-collaborative tasks, with an average performance gain of 9.93\% over the second-best result. This advantage is particularly pronounced over methods with supervised strategy modules (PPDPP) or MCTS (GDP-Zero). These results fulfill our initial motivation to develop an effective and scalable solution. Moreover, in a direct comparison with other strategy induction method (PRINCIPLES), \ours\ demonstrates an average improvement of 10.24\% across all metrics and datasets. A notable trade-off, however, is the increased dialogue length (AT) characteristic of strategy induction methods (PRINCIPLES and \ours). This highlights a potential direction for future optimization: refining the induced strategies to achieve desired outcomes with greater conversational efficiency.

\noindent\textbf{\ours~demonstrates superior practical utility in human evaluations}. We validated our findings with a human user study to measure real-world efficacy. The results, presented in Table \ref{tab:main} (right), are consistent with our simulator-based evaluations (left), with \ours\ again demonstrating the promising performance, outperforming PRINCIPLES.

\subsection{Why is \ours~Effective}
We analyze its induced strategy actions and planning logic, supplemented by ablation studies.

\subsubsection{Strategy Action Analysis}
\textbf{Setup}. We begin by characterizing the distinct properties of strategies derived from human experts, PRINCIPLES, and \ours\ through a diversity analysis. For each task, we first pool all human-crafted strategies pre-defined in the dataset with 100 randomly sampled actions each from PRINCIPLES and \ours. We then apply K-Means clustering (using \textit{bge-large-en-v1.5}) and analyze the distribution of each method's actions across these clusters. We introduce two metrics: 1) \uline{Cluster Coverage} (CC), defined as the percentage of clusters containing at least one action from the method; and 2) \uline{Within-Cluster Coverage} (WCC), defined as the average proportion and standard deviation of actions from the same method per cluster. These metrics provide a coarse- and fine-grained measure of action diversity, respectively.



\noindent\textbf{\ours~induces more diverse and nuanced actions}. Figure \ref{fig:actiond} shows results  across varying cluster numbers $K_1$ (x-axis). 1) When $K_1$=2, all methods produce fundamentally similar high-level strategy actions (CC=100\%). However, as the clusters become more fine-grained (increasing $K_1$), a clear divergence emerges. While the CC for all methods naturally declines, \ours\ consistently maintains a higher coverage, indicating that it induces a more diverse and comprehensive set of actions that span a broader range of nuanced strategic categories. 2) The low WCC for human-crafted actions is an expected results due to their limited quantity. However, we do note a slight increase in their WCC on the CB task. This occurs alongside a very low CC, suggesting that the human strategies, while semantically consistent, are concentrated within a narrow portion of the overall action space (lack of overall diversity). For both PRINCIPLES and \ours, WCC generally decreases with $K_1$, indicating their actions are distributed broadly. However, for \ours\ on the P4G dataset, its WCC increases. This indicates that \ours\ not only generates diverse actions (relatively large CC) but also creates concentrated groups of semantically similar yet subtly distinct actions within specific fine-grained clusters. To sum up, these diversities forms the foundation for our superiority.

\begin{table}[t]
\centering
\resizebox{0.49\textwidth}{!}{%
\begin{tabular}{l|l|l|l|l|l}
\toprule
\multirow{2}{*}{Method} & \multicolumn{2}{c|}{P4G} & \multicolumn{3}{c}{CB} \\  \cline{2-6}
 & SR$\uparrow$ & AT $\downarrow$ & SR$\uparrow$ & AT $\downarrow$ & SL\% $\uparrow$ \\ \midrule
\multicolumn{6}{c}{\textit{Breadth Planning Logic}} \\ \midrule
Top-1 Node & 0.760 & 4.783 & 0.465 & 5.774 & 0.140 \\
Top-3 Nodes  & 0.770 & 4.852 & 0.525 & 5.598 & 0.157 \\
Top-5 Nodes  & \textbf{0.780} & \textbf{4.756} & \textbf{0.575} & \textbf{5.383} & \textbf{0.189} \\ \midrule
\multicolumn{6}{c}{\textit{Depth Planning Logic}} \\ \midrule
Top-3 Branches & 0.760 & 4.803 & 0.485 & 5.557 & 0.150 \\
Top-2 Branches & 0.725 & 4.841 & 0.580 & 5.716 & 0.183 \\
Top-1 Branch & \textbf{0.780} & \textbf{4.756} & \textbf{0.575} & \textbf{5.383} & \textbf{0.189} \\
\midrule
1-hop Branch & 0.770& \textbf{4.636} &0.535 &5.589 &0.150\\
3-hop Branch &0.775 &4.800 &0.550 &5.491 &0.167\\
>3-hop Branch & \textbf{0.780} & 4.756 & \textbf{0.575} & \textbf{5.383} & \textbf{0.189} \\
\bottomrule
\end{tabular}%
}
\caption{Planning logic analysis.}
\label{tab:planning_diverse}
\end{table}

\begin{table*}[]
\centering
\resizebox{0.99\textwidth}{!}{%
\begin{tabular}{l|llllllllll}
\toprule
\multirow{2}{*}{\textbf{Methods}} & \multicolumn{2}{c}{\textbf{CB to P4G}} & \multicolumn{3}{c}{\textbf{P4G to CB}} & \multicolumn{2}{c}{\textbf{ALL to P4G}} & \multicolumn{3}{c}{\textbf{ALL to CB}} \\ \cmidrule(lr){2-3} \cmidrule(lr){4-6} \cmidrule(lr){7-8} \cmidrule(lr){9-11} 
 & \textbf{SR} $\uparrow$ & \textbf{AT} $\downarrow$& \textbf{SR} $\uparrow$ & \textbf{AT} $\downarrow$ & \textbf{SL\%} $\uparrow$ & \textbf{SR} $\uparrow$ & \textbf{AT} $\downarrow$& \textbf{SR} $\uparrow$ & \textbf{AT} $\downarrow$& \textbf{SL\%} $\uparrow$ \\ \midrule
Proactive \cite{deng2023prompting}&0.710 &4.59 &0.145 &\textbf{5.59} &0.138 &0.670 & \uline{4.36} &0.205 &\textbf{5.27} &0.123 \\
ProCoT \cite{deng2023prompting}&\textbf{0.755} & \textbf{4.22}&0.310 &6.24 &\textbf{0.288} &0.655 & \textbf{3.98}&0.195 &6.44 &0.138 \\
GDP-Zero \cite{yu-etal-2023-prompt}& 0.695&7.22 & 0.305& 6.31& \uline{0.254} &— &— &— &— &— \\
PPDPP \cite{deng2023plug}&0.645 &4.77 &0.225 &6.51 &0.170 &— &— &— &— &— \\
\midrule
PRINCIPLES \cite{kim-etal-2025-principles} &0.670 &5.06 &\uline{0.355}&\uline{5.66} & 0.170&\uline{0.755} &5.21 &\uline{0.325} &\uline{5.48} &\uline{0.140} \\
\ours &\textbf{0.755} &\uline{4.30} &\textbf{0.440} & 5.95&0.187 & \textbf{0.770}&4.80 & \textbf{0.540}&5.50&\textbf{0.204} \\ \bottomrule
\end{tabular}%
}
\caption{Cross-task Transferability Evaluation. \textit{CB to P4G}: Transferring CB-tailored strategies to P4G. \textit{ALL to P4G}: Transferring the union of CB and P4G strategies to P4G. Baselines without explicit strategies are omitted. `-' denotes cases omitted due to the high computational costs associated with larger strategy action sets.}
\label{tab:trans}
\end{table*}

\subsubsection{Planning Logic Analysis}
\label{plann}
\noindent\textbf{Setup}. We further analyze how breadth and depth planning logic affect \ours. For breadth, we tune the number of expended actions used for short-term suggestion generation, selecting top-k actions based on semantic similarity to current context. For depth, we leverage the multi-hop nature of branch by generating long-term suggestions from k-hop sub-branches truncated from the root. We also investigate the number of retrieval branches.



\noindent\textbf{Strategy forest equips \ours\ with enhanced strategic behavioral diversity and foresight}. While increasing the number of semantically less relevant Top-K branches provides diminishing returns, Table \ref{tab:planning_diverse} confirms that expanding the breadth (Top-k Nodes) and depth (K-hop Branch) of the planning logic is generally advantageous. This finding validates the core motivation behind our strategy forest: a structured representation of strategy actions and planning logic empowers the agent with superior strategic behavioral diversity and foresight. Thus, our default configuration employs all immediate children of the root and the top-1 full branch.


\begin{table}[]
\centering
\resizebox{0.48\textwidth}{!}{%
\begin{tabular}{l|cc|cc}
\toprule
\multicolumn{1}{c|}{\multirow{2}{*}{\begin{tabular}[c]{@{}c@{}}Strategies\end{tabular}}}	&	\multicolumn{2}{c|}{\textbf{Original Branches}}	&	\multicolumn{2}{c}{\textbf{Redundancy Removal}}	\\ \cline{2-5}
\multicolumn{1}{c|}{}	&	\multicolumn{1}{c}{URatio}	&	\multicolumn{1}{c|}{Edit Distance}	&	\multicolumn{1}{c}{URatio}	&	\multicolumn{1}{c}{Edit	Distance}	\\ \midrule
CB (Induced)	&	0.86	&	0.332	&	0.63	&	0.349	\\
P4G (Induced) &	1.00	&	0.028	&	0.98	&	0.029	\\ \midrule
CB (Origin)	&0.85	&0.322	&0.74	&0.346	\\
P4G (Origin) &1.00	&0.081	&1.00	&0.085	\\ 
\bottomrule
\end{tabular}%
}
\caption{Diversity analysis of tree branches. \textit{Induce} refers to branches with our induced actions, while \textit{Origin} uses original actions contained in the transcripts.}
\label{tab:	diverse_b}
\end{table}

\noindent\textbf{Diversity analysis on tree branches}. 
To create a representative sample, we embed all tree branches from a given task, apply K-Means clustering (K=150), and then randomly sample three branches from each cluster, yielding a corpus of 450 branches. We use two metrics: 1) \uline{Unique Ratio} (URatio), the proportion of unique branches, and 2) Average \uline{Edit Distance}, which quantifies structural similarity. Furthermore, to investigate structural redundancy, we create a variant of the corpus where consecutive repeated actions are removed, a process we term \uline{Redundancy Removal}. As shown in Table \ref{tab: diverse_b}, the strategies induced from the CB dataset are significantly less diverse than those from P4G. Analysis on Redundancy Removal setting further reveals that CB branches exhibit high structural repetition. The results of `\textit{Origin}' confirm that this is not an artifact of our induction process but rather \textit{an intrinsic limitation of the source data itself: the action sequences encoded in the original CB transcripts are not inherently diverse}. This highlights the value of transcript quality (cf. Appendix \ref{llms}). However, as evidenced by Table \ref{tab:main}, this data limitation does not diminish our overall superiority.


\begin{figure}[t]
    \centering
    \includegraphics[width=0.48\textwidth]{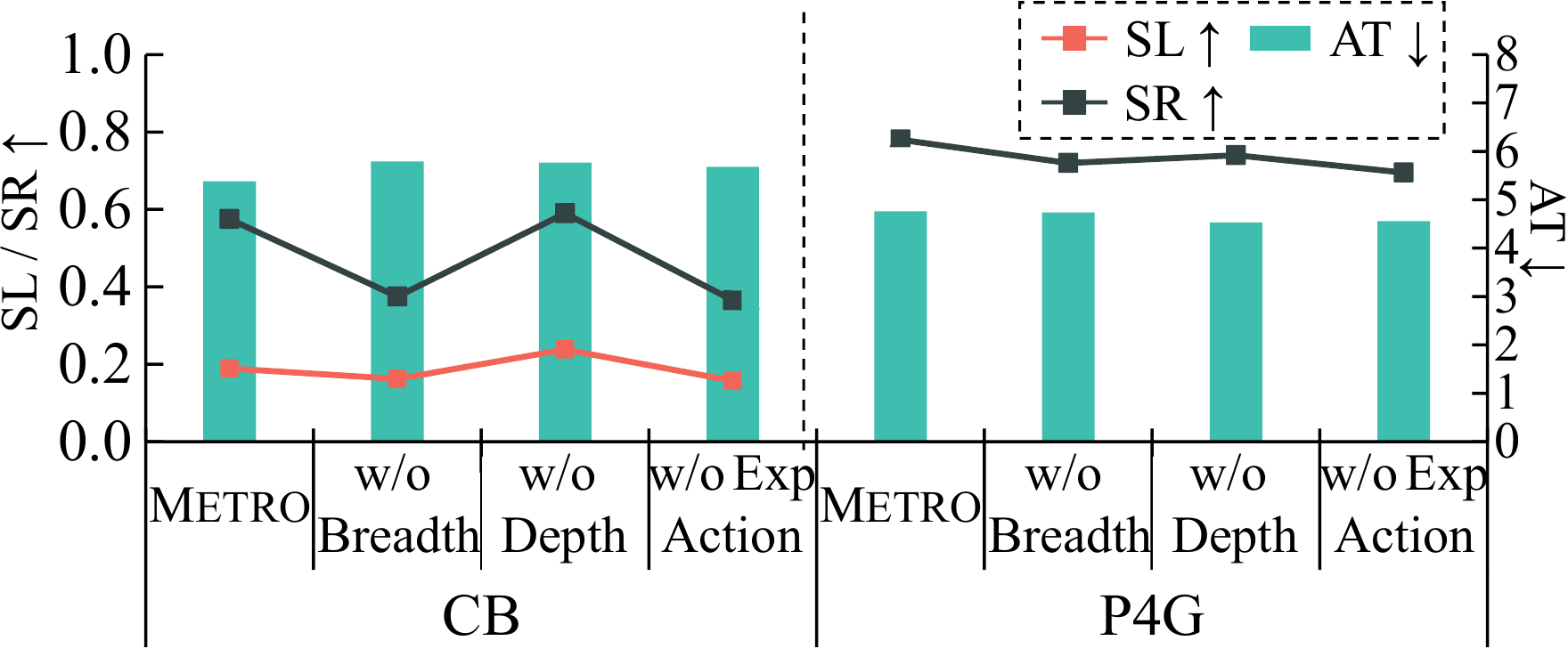}
    \caption{Ablation analysis. \textit{w/o Depth} improves performance on the CB task due to the low-diversity branches induced from the \textsc{CraigslistNegotiation} dataset.}
    \label{fig:ablation}
\end{figure}

\subsubsection{Ablation Study}
We consider the following variants. 1) \uline{w/o Exp. Action} excludes the LLM-expanded actions, using only the original actions $a_i$ for short-term suggestion generation, 2) \uline{w/o Depth} relies only on short-term suggestions, 3) \uline{w/o Breadth} uses only long-term suggestions.
According to Figure \ref{fig:ablation}, while AT remains consistent, SR and SL\% sometimes can be sensitive to the Depth component removal. On the P4G task, the ablation of any module leads to a predictable decline in performance. However, a significant exception is observed on the CB task, where the \textit{w/o Depth} configuration yields a performance enhancement. This is attributable to the inherent structural redundancy of the strategies embedded within the CB task corpus (Section \ref{plann}). This lack of diversity results in biased long-term suggestions, the removal of which is beneficial. Therefore, the efficacy of the induced strategies is contingent on the quality of the source transcripts. As elaborated in Appendix \ref{llms}, leveraging higher-quality data can further amplify our performance.



\subsubsection{Cross-Task Strategy Transferability}
\label{cross}

\noindent\textbf{\ours-induced strategies enjoy strong cross-task transferability}. As shown in Table \ref{tab:trans}, the LLM's powerful contextualization capabilities allow most methods to remain effective even when using strategies from a different task. Our analysis reveals several key insights: 1) CB strategies transfer well to P4G because they are designed for a more assertive task, but the reverse is not true, which yields weaker results; 2) The performance of PPDPP, which relies on a specialized planner, did not improve with transferred strategies after we exchanged two planners. This underscores that when the planner fails to select the right action, even a LLM, which can contextualize selected strategy,  cannot compensate for this flaw. 3) \ours\ often outperforms baselines, especially PRINCIPLES, in transfer tasks, highlighting its strategies of greater diversity and foresight. 4) We simulate strategy evolution by expanding the action space (e.g.,  CB to P4G $\to$ ALL to P4G). Most baselines suffer performance degradation, as simple LLM-based planners are prone to selection bias with a larger candidate set, while MCTS-based methods like GDP-zero incur prohibitive computational costs. In contrast, \ours\ remains effective. We attribute this to the precision of our tree retrieval mechanism, which can navigate an expanded strategy forest.


\section{Conclusion}
We pioneer an effective and scalable solution for non-collaborative dialogue by leveraging LLMs to induce strategy actions and planning logic directly from expert transcripts. We introduce \ours\ to formalize these induced strategies into a structured strategy forest. Extensive experiments demonstrate \ours's promising performance and the reason behind its success. More importantly, our work provides a investigation into tacit knowledge extraction from expert dialogues. We analyze a potential structure for representing this knowledge and explore key characteristics such as diversity and transferability. This offers insights into dialogue agent design and tacit knowledge extraction.


\section*{Limitations}
\noindent\textbf{Fidelity of Expert Knowledge}. A primary limitation of this work concerns the degree of domain expertise reflected in the source transcripts. While we utilize the \textsc{CraigslistNegotiation} and \textsc{PersuasionForGood} benchmark datasets, both of which consist of human-human dialogues, the participants are likely general users rather than professional practitioners in negotiation or persuasion. Due to the scarcity of high-fidelity transcripts produced by certified experts, it remains challenging to verify the performance of \ours\ on professional-grade expertise. Nevertheless, our empirical analysis demonstrates that inducing strategies even from general human interactions yields significant performance gains over existing baselines. We defer the evaluation of \ours\ on specialized, high-expertise corpora to future research as such data becomes available.

\noindent\textbf{In-depth Analysis on LLM-generated Transcripts}. We emphasize the value of mining domain knowledge from expert transcripts, as these records capture the tacit knowledge inherent in human interaction \cite{huang2024comatching}. Unlike explicit knowledge, this tacit knowledge is deeply embedded in individual experience and intuition \cite{ryle1945knowing, spitzer2022training}, which is difficult to convey to others through writing or verbalization~\cite{nonaka1994dynamic, lam2000tacit, gorman2002types}, and is only evident in human behaviors~\cite{dellermann201sd9hybrid}. Our method is specifically designed to formalize and utilize such tacit knowledge from unstructured behavioral data like transcripts, with specical focus on the task of non-collaborative dialogues. Therefore, the primary motivation and scope of this study is human expert behavior; we treat LLM-generated transcripts as a secondary point of comparison rather than a core research object. Although our initial findings in Appendix \ref{llms} indicate that LLM-generated transcripts can be effective, and in some cases, surpass `expert' transcripts, we leave the in-depth characterization of LLM-generated transcripts for future exploration.

\section*{Acknowledgments}
This work was supported by the National Natural Science Foundation of China (No. U25B201508, No. 62272330, and No.U24A20328).

\bibliography{custom}

\appendix

\section{Implementation Details}
\label{imp}
We conduct the main experiments on a CPU-based server with a dual-socket Intel Xeon Platinum 8352V CPU (64 physical cores, 128 threads), while the PPDPP experiments are reproduced on an NVIDIA RTX A100 GPU. 
To reduce the overall evaluation time, we adopt a mini-batch concurrent execution scheme, where evaluation samples are partitioned into small batches and processed in parallel across multiple processes. 
Dialogue state representations are obtained using the bge-large-en-v1.5 embedding model, and dialogue states are clustered with K-Means implemented in the \texttt{scikit-learn} package. 
Most experiments rely on API-based model inference using the OpenAI and Aliyun platforms. 
All implementations are based on PyTorch.For all baselines, we adhere to the implementations described in their respective papers or official GitHub repositories. Note that the first two baseline categories utilize the manually curated strategy actions provided in their original works, whereas the PRINCIPLES generates strategies autonomously. Following \citet{deng2023plug, zhang2024strength, hu2025astro}, we use GPT-3.5-turbo as the LLM backbones for conversational agents and LLM-based evaluator, respectively. To ensure a more rigorous and holistic evaluation, we also recruited human participants to interact with and assess each agent. 
Evaluation details are presented in Appendix \ref{evl}.

\subsection{Implementations of Baselines}
We implement representative LLM-based dialogue agents as baselines for comparison, adhering to their original configurations. The specific implementation details are as follows:

\noindent \textbf{Standard:} 
We simply prompt the LLM to interact with users relying solely on the provided task instructions, without employing any external dialogue strategies.

\noindent \textbf{PPDPP:} 
Following \cite{deng2023plug}, we implement the plug-and-play policy planner using RoBERTa-large \cite{liu2019roberta}, based on the original training configurations. To evaluate its performance on the P4G dataset, we adapted the task instructions and strategy descriptions to align with the persuasion scenarios. We strictly followed the original two-stage training framework (SFT and RL). For the final evaluation, we selected the model checkpoint that achieved the best performance during the testing phase.

\noindent \textbf{GDP-Zero:} 
Following \cite{yu-etal-2023-prompt}, we implement GDP-Zero using Open-Loop MCTS. We strictly adhered to the experimental settings and hyperparameters recommended in the original paper (e.g., simulation count and exploration weight). Furthermore, we extended the evaluation to the CB task by adapting the task instructions and role-playing descriptions to the negotiation context.

\noindent \textbf{Proactive:} 
Following \cite{deng2023prompting}, we implement the Proactive prompting scheme. This method requires the LLM to explicitly select a dialogue strategy from a pre-defined set before generating the response. We adapted the prompt templates to align with the specific action spaces of the P4G and CB datasets.

\noindent \textbf{ProCoT:} 
Following \cite{deng2023prompting}, we implement the Proactive Chain-of-Thought (ProCoT) framework. It instructs the LLM to generate a reasoning chain regarding dialogue status and goal planning prior to response generation. We strictly adhered to the one-shot setting and adapted the reasoning demonstrations to fit the specific negotiation and persuasion contexts.

\noindent \textbf{ICL-AIF:} 
Following \cite{fu2023improving}, we implement the ICL-AIF framework. It employs a critic model to provide natural language feedback, acting as in-context demonstrations to guide the LLM's response generation. We adapted the critic's prompts to align with the strategic goals of the P4G and CB datasets.

\noindent \textbf{PRINCIPLES:}
Following \cite{kim-etal-2025-principles}, we implement PRINCIPLES as a non-parametric strategy memory constructed via offline self-play simulations. We closely follow the original principle construction pipeline, structured principle extraction in the \emph{When–Should–Rather than–Because} format. For faithful reproduction, we adopt the principle generation procedure described in the original paper and construct a total of 296 principles for the P4G task and 100 principles for the CB task. 

\begin{algorithm}[t]
\caption{Strategy Forest Induction (Offline)}
\label{ag:al}
\begin{algorithmic}[1]
\STATE \textbf{Input}: Dialogue transcripts $\mathcal{D}$, cluster number $K$
\STATE \textbf{Output}: Strategy forest $\mathcal{F}$

\STATE \textbf{Action Extraction:} infer a data-driven action $a_i$ for each agent utterance $u_i$ in $\mathcal{D}$
\STATE \textbf{State Identification:} for each transcript $D$, build history prefixes $\{d'_i\}$ and embed states $\mathrm{Emb}(d'_i)$
\STATE \textbf{State Clustering:} run K-Means on $\{\mathrm{Emb}(d'_i)\}$ to obtain clusters and centroids $\{\mathbf{c}_k\}_{k=1}^K$
\STATE \textbf{Initialize Trees:} create one state-centric tree $T_k$ per cluster with root $\mathbf{c}_k$

\FOR{each state $d'_i$ assigned to cluster $k$}
    \STATE \textbf{Branch Generation:} extract future action path $\pi_i=(a_{i+1},\ldots)$ and insert into $T_k$ (merge identical prefixes)
    \STATE \textbf{Node Statistics:} update $n(u)$, $s(u)$ for nodes $u$ on $\pi_i$
    \STATE \textbf{Node Value:} backprop terminal outcome value to nodes with depth discount and position penalty to get $\bar{V}(u)$
\ENDFOR

\FOR{each tree $T_k$}
    \STATE \textbf{Branch Scoring:} score prefixes using Wilson lower bound $p_{\mathrm{lb}}(u)$, $\bar{V}(u)$, and $\log(1+n(u))$
    \STATE \textbf{Pruning:} use Beam Search to keep Top-$K$ highest-valued branches, prune others
\ENDFOR

\STATE \textbf{return} $\mathcal{F}=\{T_k\}_{k=1}^K$
\end{algorithmic}
\end{algorithm}

\begin{algorithm}[t]
    \caption{Inference with Strategy Forest}
    \begin{algorithmic}[1]
    \STATE \textbf{Input}: Forest $\mathcal{F}$, current dialogue history $d_{t-1}$
    \STATE Retrieve tree $f\in \mathcal{F}$ using $d_{t-1}$.
    \STATE Short-term suggestion generation $\text{Su}_s$ using immediate child nodes of the root node of $f$.
    \STATE Long-term suggestion generation $\text{Su}_l$ using highest-value branch of $f$. 
    \STATE Generate Response using $\text{Su}_s$ and $\text{Su}_l$
    \STATE \textbf{Output}: agent response at turn $t$.
    \end{algorithmic}
    \label{ag:al2}
\end{algorithm}

\subsection{Implementations of \ours}
\label{Implementations}

We conclude the pseudocode of \ours\ in Algorithm \ref{ag:al} and \ref{ag:al2}.

\subsubsection{Strategy Action Induction}
We implement strategy action induction as a two-stage pipeline over agent turns in both P4G and CB dialogues (persuader turns in P4G and buyer turns in CB). For each target turn ($\texttt{turn\_id}\ge1$), we construct a local context consisting of the preceding dialogue history, the current agent utterance, and the counterpart’s response at the same turn when available.

\paragraph{Critic.} Using \texttt{gpt-4.1-mini-2025-04-14}, we assess the target utterance under the local context and assign a label in \{\texttt{better}, \texttt{worse}, \texttt{neutral}\} with a brief rationale. We perform five independent evaluations and apply majority voting, discarding turns with fewer than three valid votes.

\paragraph{Action Expansion.}
Conditioned on the critic outcome, the same model induces a single reusable micro-principle in a fixed ``\texttt{When \dots}'' form.
Each principle is explicitly conditioned on the \emph{preceding counterpart utterance}, enabling reliable retrieval at inference time.
We induce \texttt{do} principles for \texttt{better} cases, and \texttt{avoid} principles for \texttt{worse} or \texttt{neutral} cases.

\paragraph{Example.}
In a CB dialogue at a buyer turn, suppose the seller’s previous utterance signals price firmness.
When the buyer responds by pairing a price request with additional concessions, the critic labels the turn as \texttt{better}, inducing the following micro-principle:
\medskip

\fbox{
\begin{minipage}{0.9\linewidth}
\small
\texttt{When the seller signals price firmness, do offer concrete concessions (e.g., immediate pickup or convenient payment) to improve acceptance.}
\end{minipage}
}

\medskip
In contrast, if the buyer repeats a lower price without adding new incentives, the critic assigns a \texttt{worse} label, yielding:
\medskip

\fbox{
\begin{minipage}{0.9\linewidth}
\small
\texttt{When the seller signals price firmness, avoid repeating price demands without adding new value or concessions.}
\end{minipage}
}
\medskip

These induced micro-principles are abstract strategy actions conditioned on opponent behavior and are reused during planning logic induction and inference.

\subsubsection{Dialogue Encoding and Clustering}

In the offline stage, we define a dialogue state as the dialogue history available \emph{prior to a given agent turn}, containing only past utterances.
Formally, for a transcript $D = \{u_1, u_2, \dots, u_T\}$, each candidate state corresponds to a history prefix
\[
d'_i = \{u_1, \dots, u_i\}, \quad i < T,
\]
which represents the context immediately before generating the next agent response.

\paragraph{Dialogue History Encoding.}
Each history $d'_i$ is converted into a textual sequence by concatenating all utterances in chronological order with speaker roles preserved, and encoded using the \textit{bge-large-en-v1.5} model to obtain a 1024-dimensional dense embedding.
This embedding provides a continuous semantic representation of the dialogue state.For planning logic induction, we further associate each encoded state with its corresponding \emph{future strategy trajectory}, defined as the sequence of data-driven agent actions from turn $i{+}1$ to the end of the dialogue, together with the final task outcome.

\paragraph{Dialogue State Clustering.}
All state embeddings are pooled across dialogues and optionally $L_2$-normalized, then clustered using K-Means to obtain a discrete set of dialogue state clusters.
We set the number of clusters to $K{=}150$ for P4G and $K{=}80$ for CB, reflecting differences in dialogue length and strategy diversity across tasks.
Each candidate turn is thus assigned a cluster index corresponding to the semantic neighborhood of its preceding history.

As a result, each candidate dialogue state $d'_i$ (with $i < T$) is associated with
(i) a semantic state embedding $\text{Emb}(d'_i)$,
(ii) a discrete cluster identifier,
(iii) its corresponding future action trajectory (from turn $i{+}1$ to $T$),
and (iv) the final task outcome.
These clustered states form the basis for subsequent state-centric tree induction.

\subsubsection{Planning Logic Induction}
\label{sec:planning_logic_induction_impl}

This section details the implementation of \emph{Planning Logic Induction}, which translates full expert strategy trajectories into state-conditioned planning logic represented by a strategy forest.

\paragraph{State-Centric Strategy Tree Construction and Node Aggregation.}
For each dialogue state cluster, we construct a \emph{state-centric strategy tree} to aggregate all sub-trajectories originating from that state.
The root represents the dialogue state (cluster centroid), and each sub-trajectory $(a_t,\dots,a_T)$ is inserted as a path from the root, where nodes correspond to prefixes of data-driven agent actions.
Identical action prefixes across sub-trajectories are merged, yielding a compact prefix-sharing tree that enumerates empirically observed future planning logic.Each node $u$ aggregates empirical statistics from all passing sub-trajectories, including visit count $n(u)$ and success count $s(u)$.
A sub-trajectory is successful if it reaches the task-defined terminal success state (final attitude $D$ in P4G or a successful deal in CB), yielding an empirical success probability
\[
\hat{p}(u)=\frac{s(u)}{n(u)}.
\]

Each sub-trajectory $d$ is also associated with a terminal outcome-dependent value $r(d)$ (final attitude in P4G or normalized surplus in CB), with failures assigned zero.
To reduce bias from varying dialogue lengths, we apply a relative position penalty
\[
\mathrm{pen}(d,t)=\lambda_{\mathrm{len}}\cdot\frac{t+1}{N_d},
\]
where $N_d$ is the total number of turns.
The adjusted terminal value is
\[
v(d,t)=r(d)-\mathrm{pen}(d,t).
\]

The adjusted value is propagated backward along the corresponding branch.
For a node $u$ at depth $k$ (root at $k{=}0$), the contribution from $(d,t)$ is
\[
\Delta V(u;d,t)=v(d,t)\cdot\gamma^{k}.
\]
Value signals are depth-discounted, while visit and success counts are accumulated without discounting.
The empirical average discounted value of node $u$ is
\[
\bar{V}(u)=\frac{1}{n(u)}\sum_{(d,t)\in\mathcal{I}(u)} v(d,t)\cdot\gamma^{k(u)},
\]
where $\mathcal{I}(u)$ denotes sub-trajectories passing through $u$.
Unless otherwise specified, we use $\lambda_{\mathrm{len}}=0.2$ and $\gamma=0.9$.

\paragraph{Confidence-Aware Branch Valuation and Pruning.}
High empirical performance alone can be unreliable when supported by few observations.
To mitigate overestimation of sparsely observed branches, we incorporate statistical confidence via the Wilson score lower bound $p_{\mathrm{lb}}(u)$ computed from $(s(u),n(u))$.
Each node is scored by a weighted combination of robustness, outcome quality, and empirical support:
\begin{align}
\nonumber
    S(u)&=
w_{\mathrm{sr}}\cdot p_{\mathrm{lb}}(u)
+
w_{\mathrm{val}}\cdot \bar{V}(u)
\\
\nonumber
&+
w_{\mathrm{cnt}}\cdot \log\!\big(1+n(u)\big).
\end{align}
The value of a branch is defined as the sum of its node scores along the path.
For each tree, we retain only the top-$K$ branches with the highest scores, using Beam Search for efficiency.
Unless otherwise specified, we set $w_{\mathrm{sr}}=1.0$, $w_{\mathrm{val}}=0.2$, $w_{\mathrm{cnt}}=0.05$, and $z=1.96$.

\paragraph{Breadth and Depth Planning Logic.}
After pruning, each strategy tree supports planning logic along two complementary dimensions: breadth and depth.
At inference time, we first match the current dialogue state to a dialogue state cluster, and then extract both short-term and long-term guidance from the corresponding state-centric tree.

\textbf{Breadth Planning Logic (short-term).}
Breadth logic corresponds to the \emph{immediate children of the root node}, which represent feasible one-step responses conditioned on the current dialogue state.
In our implementation, we further enrich these root-level actions by inducing and retrieving \emph{micro-principles} that abstract and reflect effective responses at this decision point.
Concretely, each micro-principle is associated with a root-level action and formulated in a fixed ``\texttt{When \dots}'' form, capturing when a particular response is appropriate.
At inference time, we embed the opponent’s most recent utterance and retrieve the top-$K$ micro-principles (we use $K{=}5$) within the matched cluster by semantic similarity.
These retrieved principles thus provide a reflective, semantically grounded representation of root-level actions, serving as short-term guidance.

\textbf{Depth Planning Logic (long-term).}
Depth logic is represented by complete root-to-leaf branches in the state-centric strategy tree, encoding long-term data-driven action sequences observed from the same dialogue state.
These branches are pre-computed offline using the confidence-aware value estimation described above.
From the matched cluster, we select the highest-ranked strategy chain (we use $k{=}1$) as the long-horizon plan.

Together, breadth logic (root-level actions enhanced by micro-principles) and depth logic (long-term strategy chains) form the strategy forest augmentation used during inference.
Rather than being applied verbatim, both forms of retrieved logic are subsequently reinterpreted and refined by the language model to adapt to the specific nuances of the current dialogue context (see Table~\ref{tab:inference_strategy_prompts}).





\section{Evaluation Details}
\label{evl}

\subsection{Persona Generation}
To improve the robustness of self-play evaluation, we follow the persona
construction protocol proposed in~\cite{zhang2024strength}.
Specifically, for both P4G and CB dialogues, we construct discrete counterpart
personas by prompting an LLM to infer the persona of the non-agent side
(the persuadee in P4G and the seller in CB) using only that participant’s
utterances.
Each persona is defined by exactly two categorical attributes: one Big-Five
personality trait and one decision-making style, resulting in a compact and
discrete specification for conditioning self-play evaluation.(see Table ~\ref {tab:ICL-AIF_p4g_persuader_prompt})

\subsection{Task Overview}
\label{sec:self_play_eval}

Following common practice in the field \cite{zhang2024strength, deng2023plug, hu2025astro}, we evaluate agents by pairing them with synthetic user simulators exhibiting diverse non-collaborative behaviors. In each interaction, both the agent and simulator alternate turns, employing strategies to maximize their respective self-interests. The dialogue proceeds until the task goal is met or the maximum number of turns (i.e.,8) is reached, after which we assess the agent's performance based on the collected interactions. Following the protocol in \citet{zhang2024strength,deng2023plug,he2025simulation}, we employ an LLM-based critic model, utilizing multi-sampling and a majority voting mechanism, to determine the termination status of each dialogue.


\paragraph{P4G.}
Persuasion outcomes are assessed by an attitude-based critic that classifies the persuadee’s final stance into four categories (\emph{reject}, \emph{neutral}, \emph{positive}, \emph{donate}), mapped to rewards $\{-1.0,-0.5,0.1,1.0\}$. We perform five parallel evaluations and aggregate by majority vote. A dialogue is considered successful if the average reward exceeds $0.6$, while outcomes below $-0.7$ are treated as clear failures; otherwise, the dialogue terminates after at most 8 turns. The critic prompt is shown in Table~\ref{tab:p4g_critic_prompt}.

\paragraph{CB.}
Negotiation outcomes are evaluated by a strict deal-based critic that determines whether an explicit agreement is reached at the end of the dialogue and extracts the agreed price if so. Success is declared only when all sampled critics agree that a deal is finalized; otherwise, the outcome is no-deal, with dialogues capped at 8 turns. The corresponding prompt is provided in Table~\ref{tab:cb_critic_prompt}.

\subsection{Interactive Evaluation}
\subsubsection{Protocol of Human Evaluation}
\label{app:human_evaluation}
We employ 5 humans with varying personas and conduct a human study in which participants interact with the agents to assess their practical effectiveness, following existing setup \cite{deng2023prompting,zhang-etal-2024-clamber,chen-etal-2024-style}. These human participants consist of adult volunteers recruited from the authors' research group and academic colleagues. Participation is entirely voluntary, and participants retain the right to withdraw at any time. They receive task instructions we delivered, and background using academic papers \cite{he2018decoupling, wang-etal-2019-persuasion}. We do not collect sensitive personal data from participants beyond their specific study responses. These are recorded without direct identifiers.

During the human evaluation phase, each participant engage in twelve distinct dialogue sessions with every evaluated method. To ensure objectivity, all assessments are conducted independently to avoid mutual interference, with final performance metrics calculated as the aggregate mean across all sessions. 

\subsubsection{Protocol of User Simulator Evaluation}

Consistent with prior research \cite{deng2023survey, bianchi2024well, huang2024conceptevaluationprotocol, zhang2024strength}, we utilize diverse LLM-based user simulators, which demonstrate non-collaborative behaviors driven by specific personas. 
Following \cite{wang-etal-2019-persuasion, jiang2024evaluating, zhang2024strength}, we incorporate two distinct persona dimensions: the \textit{Big-Five Personality} \cite{goldberg1992development} (Openness, Conscientiousness, Extraversion, Agreeableness, and Neuroticism) and the \textit{Decision-Making Styles} \cite{scott1995decision} (Directive, Conceptual, Analytical, and Behavioral), augmenting each combination with an LLM-generated description. This yields a total of 200 diverse simulators per task, distributed across the 20 distinct categories (i.e., Big-Five $\times$ Styles). Appendix \ref{usersss} shows our results on these diverse users.


\subsection{Evaluation Metrics}
\label{detailmetri}
Notably, as the evaluation metrics are calculated automatically and do not rely on subjective scoring by humans or LLMs, inter-rater evaluation reliability, whether human-human or human-LLM, is not applicable.

\subsubsection{Implementation of Automatic Evaluation Metrics}
Following \citet{deng2023plug, hu2025astro, zhang2024strength}, we consider three commonly used metrics: \underline{Success Rate} (SR), \underline{Average Turn} (AT) and \underline{Sale-to-List Ratio} (SL\%). The SR measures effectiveness by the percentage of goal achievement within a maximum number of turns, while AT measures efficiency by the average number of turns required to achieve the goal. As for the CB task, we additionally adopt the SL\% \cite{zhou2019augmenting} to determine the effectiveness of goal completion. Formally, the SL\% is expressed as $(P_{deal} - P_{target}^{seller}) / (P_{target}^{buyer} - P_{target}^{seller})$, where $P_{deal}$ is the final deal price, $P_{target}^{buyer}$ and $P_{target}^{seller}$ are the target prices of both parties. A higher SL\% represents the buyer gets more benefits from the deal. If failing to reach a deal at the end, we set SL\% as 0.



\begin{table*}[]
\centering
\resizebox{0.98\textwidth}{!}{%
\begin{tabular}{llllllllllllllll|ccc}
\toprule
\multirow{2}{*}{\textbf{Method}} & \multicolumn{3}{c}{\textbf{agreeableness}} & \multicolumn{3}{c}{\textbf{conscientiousness}} & \multicolumn{3}{c}{\textbf{extraversion}} & \multicolumn{3}{c}{\textbf{neuroticism}} & \multicolumn{3}{c}{\textbf{openness}} & \multicolumn{3}{c}{\textbf{Avg.±Std.}} \\ \cline{2-19}
 & SR & SL & AT & SR & SL & AT & SR & SL & AT & SR & SL & AT & SR & SL & AT & SR & SL & AT \\ \midrule
Standard & 0.45 & 0.37 & 5.56 & 0.03 & 0.03 & 7.00 & 0.20 & 0.16 & 5.75 & 0.03 & 0.01 & 7.00 & 0.23 & 0.21 & 6.44 & 0.19±0.18 & 0.15±0.15 & 6.35±0.68 \\
Proactive & 0.43 & 0.34 & 6.35 & 0.08 & 0.03 & 6.33 & 0.20 & 0.12 & 5.38 & 0.13 & 0.07 & 5.80 & 0.03 & 0.03 & 6.00 & 0.17±0.16 & 0.12±0.13 & 5.97±0.41 \\
ProCoT & 0.53 & 0.35 & 6.43 & 0.10 & 0.04 & 6.75 & 0.53 & 0.32 & 6.67 & 0.18 & 0.10 & 6.00 & 0.25 & 0.12 & 6.10 & 0.32±0.20 & 0.19±0.14 & 6.39±0.33 \\
ICL-AIF & 0.35 & 0.32 & 5.57 & 0.30 & 0.18 & 6.58 & 0.28 & 0.21 & 6.00 & 0.30 & 0.14 & 6.00 & 0.40 & 0.31 & 6.38 & 0.33±0.05 & \textbf{0.23±0.08} & 6.11±0.39 \\
GDPZero & 0.63 & 0.24 & 5.96 & 0.45 & 0.07 & 6.50 & 0.53 & 0.12 & 5.62 & 0.43 & 0.07 & 6.94 & 0.45 & 0.13 & 6.00 & 0.50±0.08 & 0.13±0.07 & 6.20±0.52 \\
PPDPP & 0.55 & 0.36 & 5.68 & 0.10 & 0.02 & 5.25 & 0.25 & 0.14 & 5.10 & 0.08 & 0.05 & 5.67 & 0.28 & 0.19 & 6.09 & 0.25±0.19 & 0.15±0.13 & \uline{5.56±0.39} \\
PRINCIPLES & 0.75 & 0.28 & 5.93 & 0.40 & 0.04 & 5.75 & 0.38 & 0.18 & 5.53 & 0.33 & 0.05 & 5.38 & 0.58 & 0.20 & 5.78 & \uline{0.49±0.18} & 0.15±0.10 & 5.68±0.22 \\
\ours & 0.68 & 0.31 & 5.11 & 0.45 & 0.09 & 6.00 & 0.55 & 0.21 & 5.36 & 0.60 & 0.12 & 5.21 & 0.60 & 0.21 & 5.42 & \textbf{0.58±0.08} & \uline{0.19±0.08} & \textbf{5.42±0.35} \\ \midrule\midrule
\multirow{2}{*}{\textbf{Method}} & \multicolumn{3}{c}{\textbf{analytical}} & \multicolumn{3}{c}{\textbf{behavioral}} & \multicolumn{3}{c}{\textbf{conceptual}} & \multicolumn{3}{c}{\textbf{directive}} & \multicolumn{3}{c}{\textbf{NA}} & \multicolumn{3}{c}{\textbf{Avg.±Std.}} \\ \cline{2-19}
 & SR & SL & AT & SR & SL & AT & SR & SL & AT & SR & SL & AT & & & & SR & SL & AT \\ \midrule
Standard & 0.14 & 0.10 & 5.14 & 0.20 & 0.16 & 6.20 & 0.26 & 0.25 & 5.85 & 0.14 & 0.10 & 6.29 & & & & 0.19±0.06 & 0.15±0.07 & 5.87±0.52 \\
Proactive & 0.18 & 0.11 & 5.89 & 0.32 & 0.23 & 6.38 & 0.12 & 0.11 & 5.67 & 0.06 & 0.02 & 5.33 & & & & 0.17±0.11 & 0.12±0.08 & 5.82±0.44 \\
ProCoT & 0.30 & 0.11 & 6.80 & 0.38 & 0.23 & 6.16 & 0.40 & 0.34 & 6.45 & 0.18 & 0.07 & 6.56 & & & & 0.32±0.1 & 0.19±0.12 & 6.49±0.27 \\
ICL-AIF & 0.26 & 0.17 & 6.23 & 0.50 & 0.39 & 6.28 & 0.40 & 0.29 & 6.00 & 0.14 & 0.07 & 5.57 & & & & 0.33±0.16 & \textbf{0.23±0.14} & 6.02±0.32 \\
GDPZero & 0.40 & 0.08 & 6.25 & 0.64 & 0.20 & 5.91 & 0.44 & 0.17 & 6.86 & 0.50 & 0.05 & 5.80 & & & & 0.50±0.11 & 0.13±0.07 & 6.20±0.48 \\
PPDPP & 0.10 & 0.07 & 6.00 & 0.44 & 0.33 & 5.86 & 0.26 & 0.14 & 5.46 & 0.20 & 0.06 & 5.10 & & & & 0.25±0.14 & 0.15±0.13 & \uline{5.61±0.41} \\
PRINCIPLES & 0.36 & 0.06 & 6.11 & 0.68 & 0.29 & 5.41 & 0.52 & 0.18 & 5.77 & 0.38 & 0.07 & 5.89 & & & & 0.49±0.15 & 0.15±0.11 & 5.80±0.29 \\
\ours & 0.52 & 0.14 & 5.81 & 0.76 & 0.27 & 5.53 & 0.60 & 0.26 & 5.23 & 0.42 & 0.08 & 4.81 & & & & \textbf{0.58±0.14} & 0.19±0.09 & \textbf{5.34±0.43} \\ \bottomrule
\end{tabular}%
}
\caption{Adaptability to diverse users (CB task)}
\label{tab:diverse_cb}
\end{table*}

\begin{table*}[]
\centering
\resizebox{0.98\textwidth}{!}{%
\begin{tabular}{lllllllllll|cc}
\toprule
\multirow{2}{*}{\textbf{Method}} & \multicolumn{2}{c}{\textbf{agreeableness}} & \multicolumn{2}{c}{\textbf{conscientiousness}} & \multicolumn{2}{c}{\textbf{extraversion}} & \multicolumn{2}{c}{\textbf{neuroticism}} & \multicolumn{2}{c}{\textbf{openness}} & \multicolumn{2}{c}{\textbf{Avg.±Std.}} \\ \cline{2-13}
 & SR & AT & SR & AT & SR & AT & SR & AT & SR & AT & SR & AT \\ \midrule
Standard & 0.90 & 4.31 & 0.40 & 6.69 & 0.88 & 3.26 & 0.25 & 6.10 & 0.68 & 4.74 & 0.62±0.29 & 5.02±1.38 \\
Proactive & 0.90 & 4.06 & 0.38 & 5.20 & 0.95 & 3.21 & 0.33 & 5.15 & 0.75 & 4.33 & 0.66±0.29 & \uline{4.39±0.83} \\
ProCoT & 0.93 & 4.16 & 0.43 & 5.82 & 0.93 & 3.38 & 0.35 & 6.00 & 0.78 & 3.90 & 0.68±0.28 & 4.65±1.18 \\
ICL-AIF & 0.90 & 3.56 & 0.35 & 5.36 & 0.85 & 3.38 & 0.25 & 4.80 & 0.73 & 3.90 & 0.62±0.3 & \textbf{4.2±0.85} \\
GDPZero & 0.85 & 5.29 & 0.25 & 5.70 & 0.88 & 4.80 & 0.65 & 6.15 & 0.68 & 5.22 & 0.66±0.25 & 5.43±0.51 \\
PPDPP & 0.98 & 4.56 & 0.45 & 5.89 & 0.95 & 3.74 & 0.48 & 5.89 & 0.80 & 4.50 & 0.73±0.25 & 4.92±0.95 \\
PRINCIPLES & 0.95 & 4.87 & 0.50 & 5.80 & 0.95 & 4.34 & 0.55 & 6.32 & 0.90 & 5.61 & \uline{0.77±0.23} & 5.39±0.78 \\
OURS & 0.98 & 4.54 & 0.65 & 5.77 & 1.00 & 4.08 & 0.45 & 5.11 & 0.83 & 4.85 & \textbf{0.78±0.23} & 4.87±0.63 \\ \midrule\midrule
\multirow{2}{*}{\textbf{Method}} & \multicolumn{2}{c}{\textbf{analytical}} & \multicolumn{2}{c}{\textbf{behavioral}} & \multicolumn{2}{c}{\textbf{conceptual}} & \multicolumn{2}{c}{\textbf{directive}} & \multicolumn{2}{c}{\textbf{NA}} & \multicolumn{2}{c}{\textbf{Avg.±Std.}} \\ \cline{2-13}
 & SR & AT & SR & AT & SR & AT & SR & AT & & & SR & AT \\ \midrule
Standard & 0.38 & 5.89 & 0.78 & 4.54 & 0.70 & 4.66 & 0.62 & 3.65 & & & 0.62±0.17 & 4.68±0.93 \\
Proactive & 0.38 & 5.47 & 0.72 & 4.28 & 0.80 & 4.03 & 0.74 & 3.35 & & & 0.66±0.19 & \uline{4.28±0.89} \\
ProCoT & 0.42 & 5.76 & 0.68 & 3.50 & 0.84 & 4.81 & 0.78 & 3.62 & & & 0.68±0.19 & 4.42±1.07 \\
ICL-AIF & 0.32 & 5.38 & 0.72 & 3.67 & 0.76 & 4.00 & 0.66 & 3.30 & & & 0.62±0.2 & \textbf{4.09±0.91} \\
GDPZero & 0.32 & 6.50 & 0.84 & 5.07 & 0.80 & 5.53 & 0.68 & 4.94 & & & 0.66±0.24 & 5.51±0.71 \\
PPDPP & 0.56 & 5.79 & 0.84 & 4.43 & 0.84 & 4.79 & 0.68 & 3.91 & & & 0.73±0.14 & 4.73±0.79 \\
PRINCIPLES & 0.72 & 6.28 & 0.80 & 4.70 & 0.82 & 5.22 & 0.74 & 4.84 & & & \uline{0.77±0.05} & 5.26±0.71 \\
OURS & 0.72 & 5.78 & 0.80 & 4.38 & 0.86 & 4.74 & 0.74 & 4.19 & & & \textbf{0.78±0.06} & 4.77±0.71 \\ \bottomrule
\end{tabular}%
}
\caption{Adaptability to diverse users (P4G task)}
\label{tab:diverse_p4g}
\end{table*}

\section{Additional Experimental Results}
\label{moreex}
\subsection{Additional Analysis on Human Study}
\ours~demonstrates superior practical utility in human evaluations. We validated our findings with a human user study to measure real-world efficacy. The results, presented in Table \ref{tab:main} (right), are consistent with our simulator-based evaluations (left), with \ours\ again demonstrating the promising performance. Qualitative feedback from participants revealed that while most baseline models exhibited strategic rigidity, often relying on repetitive, formulaic requests for donations (e.g., `\textit{Your donation would go towards providing essential resources like [xx]]}'), \ours\ demonstrates significantly higher conversational flexibility. For instance, whereas other models frequently reached an impasse when a donation was refused, \ours\ is capable of strategic pivoting, such as suggesting social media advocacy as an alternative: `\textit{I completely understand. If donating isn't feasible right now, would you consider sharing Save the Children on social media to help raise awareness instead?}'


\subsection{Performance on Various User Simulators}
\label{usersss}
\noindent\textbf{\ours~exhibits superior adaptability to diverse users}. We evaluate assess the generalization capacity of our induced strategies across diverse user types. To this end, we categorize user simulators into 9 distinct groups based on the Big Five Personality traits and their Decision-Making Styles. The results, depicted in Table \ref{tab:diverse_cb} and Table \ref{tab:diverse_p4g}, indicate the robust superiority of \ours\ across all user types. Basically, since most baselines rely on a fixed, pre-defined set of strategy actions, they lack the adaptability to tailor their approach to different users. This limitation is partially mitigated in PRINCIPLES, which also leverages induced strategies, allowing for greater flexibility. However, \ours\ still maintains a consistent advantage, and the underlying mechanisms driving this superiority are detailed in the subsequent sections.


\begin{table}[]
\centering
\resizebox{0.4\textwidth}{!}{%
\begin{tabular}{l|l|l|l}
\toprule
\multirow{2}{*}{Transcript Type} &\multicolumn{3}{c}{CB} \\  \cline{2-4}
  & SR$\uparrow$ & AT $\downarrow$ & SL\% $\uparrow$ \\ \midrule
 Expert &0.440 &5.977 &0.166 \\
Non-Expert &0.315 &5.762 &0.131 \\
LLM-Generated &0.500&5.790 &0.149 \\
\bottomrule
\end{tabular}%
}
\caption{Transcripts Analysis}
\label{tab:LLM_trans}
\end{table}

\subsection{Impact of Transcript Quality}
\label{llms}
While expert transcripts are valuable, they are often limited in inconsistent quality. To assess the potential of \ours, we analyze the following three types of transcripts using CB task: 1)\uline{Expert Transcript}.We manually select 500 high-quality dialogues from the \textsc{CraigslistBargain} dataset \cite{he2018decoupling} to approximate effective human negotiation behavior. Specifically, selected dialogues are required to (i) contain at least several rounds of bidirectional interaction with explicit counter-offers; (ii) exhibit incremental and directionally consistent price movements, where concessions or rejections are clearly grounded in the dialogue context (e.g., item condition, payment method, delivery constraints, or timing); and (iii) reflect stable and interpretable negotiation patterns, such as anchoring followed by gradual concession or conditional trade-offs.
2)\uline{Non-Expert Transcript}.
In contrast, we select 500 low-quality dialogues from the same dataset to simulate suboptimal human performance. These dialogues typically involve truncated or weakly interactive structures, irregular or abrupt price changes without contextual justification, or fragmented exchanges in which negotiation strategies are either absent or inconsistent across turns, making them less informative for learning transferable bargaining behavior. 3) \uline{LLM-Generated Transcript}. We investigate the utility of LLM-generated dialogues to determine if it can serve as a viable alternative to human data. Here, we refer to GPT-5 to generate 500 high-quality dialogues simulating the CB task.

Table \ref{tab:LLM_trans} clearly shows that transcript quality is critical (Expert > Non-Expert), with LLM-generated data surprisingly yielding the best performance. We posit this is not because LLMs surpass human experts, but because our 'Expert' data, which was sourced from AMT according to \citet{he2018decoupling}, likely does not represent true domain expertise (i.e., the participants are likely general users rather than professional practitioners in negotiation or persuasion). This leads to two robust conclusions: first, the quality of transcripts is paramount for effective strategy induction, and second, LLM-generated transcripts can serve as a potent and scalable substitute for human data when expert collection is infeasible.





\subsection{Efficiency Analysis} Unlike methods such as GDP-Zero that rely on computationally expensive test-time MCTS inference, our approach is designed for high efficiency. The strategy forest is pre-computed offline. Before generating a response, \ours\ only needs to perform a single cosine similarity search to retrieve the relevant strategy tree. The LLM backbone then contextualizes the retrieved short- and long-term strategies to generate suggestions for the current turn. This design results in minimal computational overhead during runtime, with the only addition being a single extra LLM prompt for suggestion generation per turn. In terms of storage, while the strategy forest does introduce a memory footprint, it remains modest. The primary storage cost comes from the root node embeddings. In our experiments, with N=150 trees, the entire forest required only $\sim$25 MB of storage space, which is negligible on modern hardware. In summary, our method is not only highly effective but also demonstrates excellent practical utility, combining low-latency inference with a minimal storage footprint.

\section{Case Studies}
\begin{figure*}[ht]
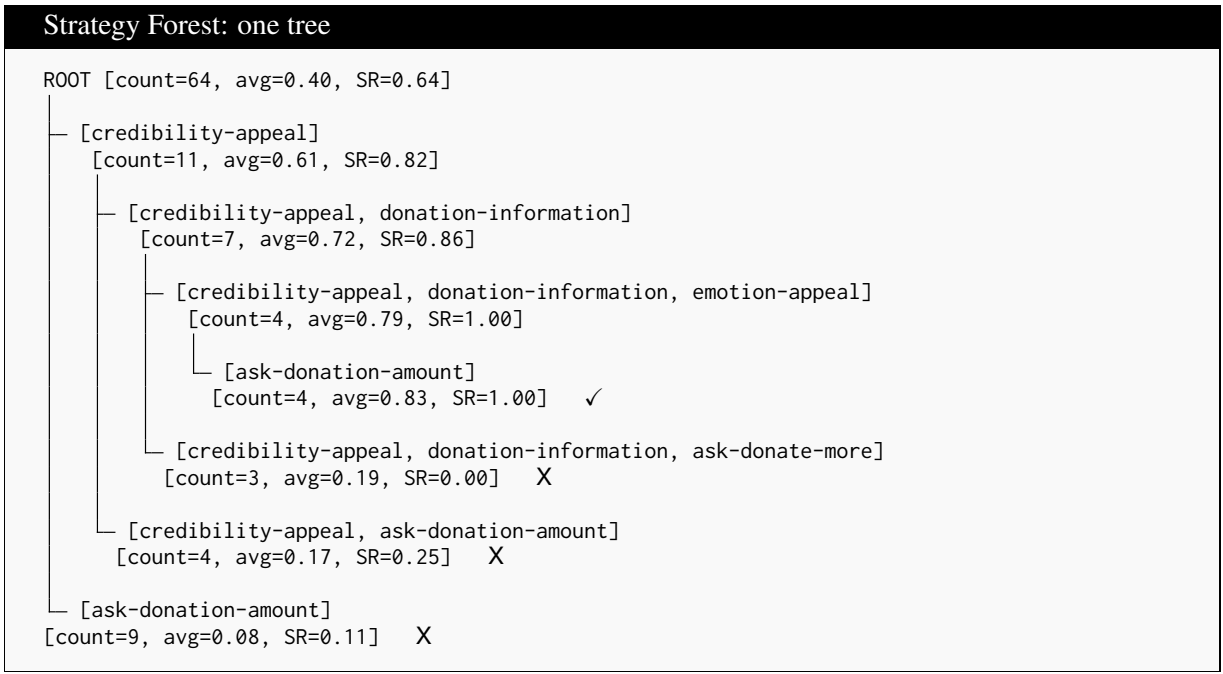
 
    \centering
    \begin{tcolorbox}[
        colback=gray!5!white, 
    colframe=black!75!black, 
    title=Strategy Forest: one tree,
    fontupper=\ttfamily\small, 
    sharp corners,
    boxrule=0.5pt,
    width=\linewidth
    ]
    ROOT [count=64, avg=0.40, SR=0.64] \\
│ \\
├─ [credibility-appeal] \\
│ \quad [count=11, avg=0.61, SR=0.82] \\
│ \quad │ \\
│ \quad ├─ [credibility-appeal, donation-information] \\
│ \quad │ \quad [count=7, avg=0.72, SR=0.86] \\
│ \quad │ \quad │ \\
│ \quad │ \quad ├─ [credibility-appeal, donation-information, emotion-appeal] \\
│ \quad │ \quad │ \quad [count=4, avg=0.79, SR=1.00] \\
│ \quad │ \quad │ \quad │ \\
│ \quad │ \quad │ \quad └─ [ask-donation-amount] \\
│ \quad │ \quad │ \quad \quad [count=4, avg=0.83, SR=1.00] \quad \checkmark \\
│ \quad │ \quad │ \\
│ \quad │ \quad └─ [credibility-appeal, donation-information, ask-donate-more] \\
│ \quad │ \quad \quad [count=3, avg=0.19, SR=0.00] \quad \textsf{X} \\
│ \quad │ \\
│ \quad └─ [credibility-appeal, ask-donation-amount] \\
│ \quad \quad [count=4, avg=0.17, SR=0.25] \quad \textsf{X} \\
│ \\
└─ [ask-donation-amount] \\
\quad [count=9, avg=0.08, SR=0.11] \quad \textsf{X}
    \end{tcolorbox}
    \caption{Induced Strategy Tree in the P4G dialog} 
    \label{fig:strategy_tree} 
\end{figure*}

For better understanding our methods, we provide illustrative cases of the induced Strategy Forest and resulting dialogue sessions produced by \ours. Specifically, Table \ref{tab:p4g_strategies} and Table \ref{tab:trip_strategies} detail human-crafted strategies (actions, in particular), while Table \ref{tab:baseline_principles_blocks} provides the PRINCIPLES-induced strategies. In contrast, \ours-induced strategy forest is presented in Table \ref{tab:ours_metro_guidance_and_hints} and Figure \ref{fig:strategy_tree}. 
As illustrated, the \ours-induced strategies possess a distinct hierarchical structure. Unlike the flat, independent lists of actions found in baseline methods, this architecture organizes individual strategy actions into coherent, multi-turn trajectories, effectively mapping out the planning logic required to bridge the semantic gap between initial user inquiries and the target information.

Additionally, we illustrate the conversation cases of \ours\ and PRINCIPLES in Table \ref{tab:case_ours_metro} and Table \ref{tab:case_baseline_principles}, respectively. Basically, \ours\ demonstrates superior strategic pacing and contextual adaptability compared to the PRINCIPLES. Based on the case study, While PRINCIPLES becomes preoccupied with excessive rapport-building, inducing user impatience and conversational friction, \ours effectively bridges the semantic gap by transitioning logically from greetings to the core task. As seen in Table \ref{tab:case_baseline_principles}, the PRINCIPLES ignores explicit user signals (e.g., being in a hurry) and adheres to a rigid script, whereas \ours\ (Table \ref{tab:case_ours_metro}) maintains a collaborative dialogue, addressing specific user inquiries while steadily advancing toward the donation goal. These case studies empirically validate that \ours-induced strategies provide more than just short-term response accuracy; they provide structural foresight. By organizing expert knowledge into a Strategy Forest, the agent gains the ability to navigate complex multi-turn transitions. This hierarchical representation allows the agent to maintain contextual alignment (Breadth) while ensuring goal-oriented progress (branch-level Depth), effectively mitigating the strategic rigidity and user friction observed in baseline methods


\definecolor{persuadeegray}{gray}{0.45}

\begin{table*}[t]
\centering
\small
\setlength{\tabcolsep}{4pt}
\begin{tabularx}{\linewidth}{c l X}
\toprule
\textbf{Turn} & \textbf{Speaker} & \textbf{Utterance} \\
\midrule
0 & \textbf{Persuader} & \textbf{Hey. How are you?} \\
0 & \textcolor{persuadeegray}{Persuadee} & \textcolor{persuadeegray}{I'm well, how are you?} \\ \\
1 & \textbf{Persuader} & \textbf{What have you been up to lately?} \\
1 & \textcolor{persuadeegray}{Persuadee} & \textcolor{persuadeegray}{not much? what about you} \\ \\
2 & \textbf{Persuader} & \textbf{I've been involved in supporting Save the Children, and I recently heard a touching story about how even a small donation can make a big difference in a child's life.} \\
2 & \textcolor{persuadeegray}{Persuadee} & \textcolor{persuadeegray}{yeah? can you tell me about it?} \\ \\
3 & \textbf{Persuader} & \textbf{Absolutely, the story I heard was about a child who received lifesaving medical treatment thanks to donations to Save the Children.} \\
3 & \textcolor{persuadeegray}{Persuadee} & \textcolor{persuadeegray}{all right, what is the charity group's duty} \\ \\
4 & \textbf{Persuader} & \textbf{Save the Children's duty is to provide children with education, healthcare, and protection to ensure a better future for them.} \\
4 & \textcolor{persuadeegray}{Persuadee} & \textcolor{persuadeegray}{sounds good, what do you expect me to do} \\ \\
5 & \textbf{Persuader} & \textbf{Would you consider making a small donation to Save the Children to help provide essential support to children in need?} \\
5 & \textcolor{persuadeegray}{Persuadee} & \textcolor{persuadeegray}{yes, how much should I donate} \\ \\
6 & \textbf{Persuader} & \textbf{That's wonderful! Any amount you feel comfortable with will make a meaningful difference to children in need.} \\
6 & \textcolor{persuadeegray}{Persuadee} & \textcolor{persuadeegray}{0.7\$? is that ok?} \\  \\
7 & \textbf{Persuader} & \textbf{Thank you for your generosity! Your \$0.7 donation will truly make a difference in a child's life.} \\
7 & \textcolor{persuadeegray}{Persuadee} & \textcolor{persuadeegray}{ok I will donate 0.7\$} \\ 
\bottomrule
\end{tabularx}
\caption{Case study (METRO). }
\label{tab:case_ours_metro}
\end{table*}

\begin{table*}[t]
\centering
\small
\setlength{\tabcolsep}{4pt}
\begin{tabularx}{\linewidth}{c l X}
\toprule
\textbf{Turn} & \textbf{Speaker} & \textbf{Utterance} \\
\midrule
0 & \textbf{Persuader} & \textbf{Hi} \\
0 & \textcolor{persuadeegray}{Persuadee} & \textcolor{persuadeegray}{Hello, how are you this evening?} \\ \\
1 & \textbf{Persuader} & \textbf{I'm doing well, thank you. How has your day been so far?} \\ 
1 & \textcolor{persuadeegray}{Persuadee} & \textcolor{persuadeegray}{a wonderful day!} \\ \\
2 & \textbf{Persuader} & \textbf{That's great to hear! What was the highlight of your wonderful day?} \\
2 & \textcolor{persuadeegray}{Persuadee} & \textcolor{persuadeegray}{i got prompted by my boss} \\ \\
3 & \textbf{Persuader} & \textbf{That's fantastic! What did your boss praise you for?} \\
3 & \textcolor{persuadeegray}{Persuadee} & \textcolor{persuadeegray}{hahaha! i guess, thank you for that, by the way, what do you want} \\ \\
4 & \textbf{Persuader} & \textbf{I'm actually here to talk to you about an opportunity to make a difference with a small donation to Save the Children.} \\
4 & \textcolor{persuadeegray}{Persuadee} & \textcolor{persuadeegray}{ok bye} \\ \\
5 & \textbf{Persuader} & \textbf{Before you head off, I'd love to share a heartwarming story about how a \$2 donation helped provide a child with life-saving medical treatment.} \\
5 & \textcolor{persuadeegray}{Persuadee} & \textcolor{persuadeegray}{quickly, im in a hurry} \\ \\
6 & \textbf{Persuader} & \textbf{With just a \$2 donation, a child received life-saving medical treatment and is now healthy and thriving.} \\
6 & \textcolor{persuadeegray}{Persuadee} & \textcolor{persuadeegray}{here is 0.5\$} \\ \\
7 & \textcolor{persuadeegray}{Persuadee} & \textcolor{persuadeegray}{i have donated, bye} \\
\bottomrule
\end{tabularx}
\caption{Case study (Baseline: PRINCIPLES).}
\label{tab:case_baseline_principles}
\end{table*}

\begin{table*}[t]
\centering
\small
\setlength{\tabcolsep}{4pt}
\renewcommand{\arraystretch}{1.15}
\begin{tabularx}{\linewidth}{c X}
\toprule
\textbf{Turn} & \textbf{PRINCIPLES strategy} \\
\midrule

1 & ……\\

2 &
\textbf{(1)} When the persuadee shares positive news about their day, you should acknowledge their good mood and continue the conversation by asking about the highlight of their day rather than abruptly changing the subject because this approach shows genuine interest in their well-being and maintains a positive tone in the interaction. \newline
\textbf{(2)} When the persuadee asks about their well-being, you should reciprocate by acknowledging their positive sentiment and continue the conversation by expressing genuine interest in their day, as this helps in building rapport and fostering a positive interaction. \newline
\textbf{(3)} When the persuadee expresses positive emotions or experiences, you should acknowledge and validate their feelings by responding with enthusiasm and positivity because this helps establish a friendly and supportive atmosphere for the conversation. \\

3 &……\\

4 &……\\

5 &…… \\

6 &
\textbf{(1)} When the Persuadee hesitates to share specific examples of their successes, you should acknowledge the complexity of the situation and shift the conversation towards discussing their overall strengths and qualities instead of pressing for detailed examples. This approach helps maintain a positive and supportive dialogue. \newline
\textbf{(2)} When the persuadee struggles to provide specific examples of their work achievements, you should acknowledge the complexity of the situation and shift the focus to discussing their overall dedication and work ethic. This approach validates the persuadee's efforts and encourages them to reflect on their strengths in a positive light. \newline
\textbf{(3)} When the persuadee struggles to provide a specific example of their dedication making a difference, you should acknowledge the complexity of the situation and shift the focus towards discussing their overall work ethic and the qualities that contribute to their success, rather than pressing for a particular instance, because this shows understanding and encourages reflection on their strengths. \\

7 &…… \\

8 &
\textbf{(1)} When the persuadee expresses interest in hearing more about the opportunity you are presenting, you should provide compelling details about the positive impact of the donation to Save the Children rather than focusing solely on the process of donation collection. This approach builds enthusiasm and addresses the persuadee's curiosity more effectively. \newline
\textbf{(2)} When the persuadee expresses interest in hearing more about the opportunity to make a difference with a small donation to Save the Children, you should share a compelling success story that highlights the impact of donations, backed by relevant statistics and details, rather than solely discussing the general benefits of the contribution, as this showcases tangible outcomes and aligns more closely with the persuadee's interests. \newline
\textbf{(3)} When the persuadee shows interest in the opportunity to make a difference through a donation, you should provide engaging details about the specific projects or initiatives supported by Save the Children. This helps create a personal connection and allows the persuadee to visualize the impact of their contribution, fostering a deeper understanding of how their donation can create positive change. \\

\bottomrule
\end{tabularx}
\caption{Strategies for  \textsc{PRINCIPLES} method in the P4G task}
\label{tab:baseline_principles_blocks}
\end{table*}

\begin{table*}[t]
\centering
\small
\setlength{\tabcolsep}{4pt}
\renewcommand{\arraystretch}{1.15}
\begin{tabularx}{\linewidth}{c X X}
\toprule
\textbf{Turn} & \textbf{Bread guidance} & \textbf{Depth guidance} \\
\midrule

1 &
Clearly and succinctly explain the charity's mission and impact to generate enthusiasm and encourage further engagement from the persuadee. &
Given the persuadee's lack of familiarity with the charity, the persuader should focus on establishing credibility, gradually introducing logical appeals, and ultimately guiding the persuadee towards a proposition of donation through continuous credibility building and gratitude, ensuring a positive and logical reinforcement of the cause. \\

2 &……&……\\

3 &……&……\\

4 &
Clearly outline the specific impact their donation will make and provide a suggested donation amount to encourage immediate contribution. &
Given the persuadee's initial lack of clarity on the desired action, the persuader should aim to gradually build credibility, emotional connection, and personal relevance in their messaging to ultimately lead the persuadee to feel motivated and empowered to make a donation as a meaningful and self-driven choice. \\

5 &……&……\\

6 &……&……\\

7 &
Reinforce the positive intention and impact of their donation to encourage deeper commitment before ending the conversation. &
Given the abrupt end to the conversation, the persuader should focus on building a stronger emotional connection with the persuadee by sharing more personal stories and increasing credibility over time to foster a sense of trust and commitment towards the cause. \\

8 &……&……\\

\bottomrule
\end{tabularx}
\caption{Extracted strategies for \textsc{METRO} method in the P4G task}
\label{tab:ours_metro_guidance_and_hints}
\end{table*}


\section{Prompts}
We details all prompts used in our experiments.

\begin{table*}[h!]
    \centering
    \renewcommand{\arraystretch}{1.2}
    \begin{tabular}{|p{0.95\linewidth}|}
    \hline
    \textbf{Prompts for the METRO Buyer model in the CB task} \\
    \hline

    \textbf{[System]} \\
    You are a Buyer negotiating the price of a $[$Item Name$]$. \\
    Your target price is $[$Buyer Target Price$]$. \\
    Reply in 1--2 natural sentences. \\
    \\
    \textbf{[User]} \\
    Item description: $[$Item Description$]$ \\
    Conversation so far: $[$Conversation History$]$ \\

    This summarizes successful bargaining trajectories in similar dialogues: 
    $[$Strategy Chain Hint Text$]$ \\

    This captures the most effective response to the seller's latest message: $[$High Level Guidance$]$ \\

    \texttt{[Instruction]} \\
    Think from two perspectives: \\
    1) Long-term planning --- use the long-term planning summary to maintain a good bargaining trajectory. \\
    2) Immediate response --- use the high-level guidance to respond appropriately to the seller's latest message. \\
    Write the Buyer's next reply in 1--2 short, natural sentences only. \\
    \hline
    \end{tabular}
    \caption{Prompts for the METRO Buyer model in the CB task}
    \label{tab:metro_cb_buyer_prompt}
\end{table*}

\begin{table*}[h!]
    \centering
    \renewcommand{\arraystretch}{1.2}
    \begin{tabular}{|p{0.95\linewidth}|}
    \hline
    \textbf{Prompts for the METRO Persuader model in the P4G task} \\
    \hline

    \textbf{[System]} \\
    You are the Persuader in a persuasion conversation. \\
    \textbf{Ultimate Goal:} \\
    - Guide the Persuadee to make a donation to Save the Children as soon as it becomes appropriate. \\
    - The donation does not need to be large; even \$1--\$2 meaningfully helps children in need. \\
    - Avoid excessive questioning that stalls progress; questions should move closer to donation. \\
    \\
    \textbf{[User]} \\
    Conversation : $[$Conversation History$]$ \\
    This summarizes successful bargaining trajectories in similar dialogues:  $[$Strategy Chain Hint Text$]$ \\

   This captures the most effective response to the persuadee's latest message:  $[$High Level Guidance$]$ \\

    \texttt{[Instruction]} \\
    Think from two perspectives: \\
    1. Long-term planning --- use the long-term planning summary to understand the overall strategic direction and key phases of persuasion. \\
    2. Immediate response --- use the high-level guidance to adapt sensitively to the user's latest message. \\
    Combine both perspectives to decide your next move. \\
    Write the Persuader's next reply in 1--2 short, natural sentences only. \\
    \hline
    \end{tabular}
    \caption{Prompts for the METRO Persuader model in the P4G task}
    \label{tab:metro_p4g_persuader_prompt}
\end{table*}

\begin{table*}[h!]
    \centering
    \renewcommand{\arraystretch}{1.2}
    \begin{tabular}{|p{0.95\linewidth}|}
    \hline
    \textbf{The comprehensive prompt of user simulators in the CB task.} \\
    \hline
    \textbf{[System]} \\
    You are the Seller in a price bargaining game. \\
    Persona: $[$Persona Description$]$ \\
    You are selling a $[$Item Name$]$. \\
    Your target price is $[$Seller Target Price$]$. \\
    You prefer to sell this item at this price. \\
    Style rules: \\
    - Reply in 1--2 sentences. \\
    - Your utterances and bargain behavior need to strictly follow your persona. Varying your wording and avoid repeating yourself verbatim. \\
    - You can decide to change your target price flexibly based on your persona and the conversation. \\
    \hline
    \textbf{[User]} \\
    Item description: $[$Item Description$]$ \\
    Conversation so far:  $[$Conversation History$]$ \\
    Write the Seller's next reply. \\
    \hline
    \end{tabular}
    \caption{The comprehensive prompt of user simulators in the CB task.}
    \label{tab:cb_seller_persona_prompt}
\end{table*}

\begin{table*}[h!]
    \centering
    \renewcommand{\arraystretch}{1.2}
    \begin{tabular}{|p{0.95\linewidth}|}
    \hline
    \textbf{The comprehensive prompt of user simulators in the P4G task.} \\
    \hline
    \textbf{[System]} \\
    You are now entering role-playing mode. \\
    You are the Persuadee in a persuasion dialogue. \\
    Act according to the following personality description: \\
    {[}Persona Description{]} \\
    Important: \\
    - You reply based on your personality, your emotions, and your thinking style. \\
    - Your willingness for donation depends on your persona and how influenced you are by the Persuader. \\
    - You need to determine whether to donate money or not. If you think your are convinced by the Persuader, you should donate money for the charity. \\
    \hline
    \textbf{[User]} \\
    conversation:  $[$Conversation History$]$ \\
    Please reply with ONE short sentence. \\
    \hline
    \end{tabular}
    \caption{The comprehensive prompt of user simulators in the P4G task.}
    \label{tab:p4g_Persuadee_persona_prompt}
\end{table*}

\begin{table*}[h!]
    \centering
    \renewcommand{\arraystretch}{1.3}
    \begin{tabularx}{\linewidth}{l X}
    \toprule
    \textbf{Dialogue Strategy} & \textbf{Natural Language Form} \\
    \midrule
    Logical appeal & Please use of reasoning and evidence to convince the persuadee. \\
    Emotion appeal & Please elicit the specific emotions to influence the persuadee. \\
    Credibility appeal & Please use credentials and cite organizational impacts to establish credibility and earn the user's trust. The information usually comes from an objective source (e.g., the organization's website or other well-established websites). \\
    Task-related inquiry & Please ask about the persuadee opinion and expectation related to the task, such as their interests in knowing more about the organization. \\
    Source-related inquiry & Please ask if the persuadee is aware of the organization (i.e., the source in our specific donation task). \\
    Personal-related inquiry & Please ask about the persuadee previous personal experiences relevant to charity donation. \\
    Donation information & Please provide specific information about the donation task, such as the donation procedure, donation range, etc. By providing detailed action guidance, this strategy can enhance the persuadee's self-efficacy and facilitates behavior compliance. \\
    Personal story & Please use narrative exemplars to illustrate someone donation experiences or the beneficiaries positive outcomes, which can motivate others to follow the actions. \\
    Self-modeling & Please use the self-modeling strategy where you first indicate the persuadee own intention to donate and choose to act as a role model for the persuadee to follow. \\
    Foot in the door & Please use the strategy of starting with small donation requests to facilitate compliance followed by larger requests. \\
    \bottomrule
    \end{tabularx}
    \caption{Strategy actions of P4G and their corresponding natural language prompt}
    \label{tab:p4g_strategies}
\end{table*}

\begin{table*}[h!]
    \centering
    \renewcommand{\arraystretch}{1.3}
    \begin{tabularx}{\linewidth}{l X}
    \toprule
    \textbf{Dialogue Strategy} & \textbf{Natural Language Form} \\
    \midrule
    Greetings & Please say hello or chat randomly. \\
    Ask a question & Please ask any question about product, year, price, usage, etc. \\
    Answer a question & Please provide information about the product, year, usage, etc. \\
    Propose the first price & Please initiate a price or a price range for the product. \\
    Propose a counter price & Please propose a new price or a new price range. \\
    Use comparatives & Please propose a vague price by using comparatives with existing price. \\
    Confirm information & Please ask a question about the information to be confirmed. \\
    Affirm confirmation & Please give an affirmative response to a confirm. \\
    Deny confirmation & Please give a negative response to a confirm. \\
    Agree with the proposal & Please agree with the proposed price. \\
    Disagree with a proposal & Please disagree with the proposed price. \\
    \bottomrule
    \end{tabularx}
    \caption{Strategy actions of CB and their corresponding natural language prompt}
    \label{tab:trip_strategies}
\end{table*}

\begin{table*}[h!]
    \centering
    \renewcommand{\arraystretch}{1.2}
    \begin{tabular}{|p{0.95\linewidth}|}
    \hline
    \textbf{Prompts for critic model in P4G task} \\
    \hline
    \textbf{[System]} \\
    Given a conversation between a Persuader and a Persuadee, please assess whether the Persuadee has agreed to donate to Save the Children. \\
    \hline
    \textbf{[User]} \\
    The following is a conversation: $[$Conversation History$]$ \\
    Question: Has the Persuadee agreed to donate to Save the Children? \\
    You can only answer using the following sentences: \\
    A) The persuadee has explicitly refused. \\
    B) The persuadee remains neutral about donating. \\
    C) The persuadee has a positive attitude towards donating but hasn't decided yet. \\
    D) The persuadee has decided to donate. \\
    Answer (one letter only): \\
    \hline
    \end{tabular}
    \caption{Prompts for critic model in P4G task}
    \label{tab:p4g_critic_prompt}
\end{table*}

\begin{table*}[h!]
    \centering
    \renewcommand{\arraystretch}{1.0}
    
    \begin{tabular}{|p{0.95\linewidth}|}
    \hline
    \textbf{Prompts for critic model in CB task} \\
    \hline
    \textbf{[System]} \\
    You are a strict judge for a price negotiation between a Buyer and a Seller. 
    Your task: decide whether a deal is reached AT THE END of the conversation. \\
    IMPORTANT: Use a STRICT definition of "deal": \\
    - deal=true ONLY IF there is an explicit acceptance/confirmation to finalize the deal. \\
    - The Buyer must clearly accept a specific price (a number) or explicitly accept the Seller's last stated price. \\
    - If the conversation ends with a counteroffer, a final offer, a question, negotiation, or "does that work?", then deal=false. \\
    Price extraction rules: \\
    - If deal=true, output the agreed numeric price. \\
    - If multiple prices appear, choose the one that is explicitly accepted in the final agreement. \\
    - If deal=false, price must be null. \\
    Output STRICT JSON only: \{"deal": true/false, "price": number or null\} \\
    Do not output any other text. \\
    \hline
    \textbf{[User]} \\
    Decide if a deal is reached at the end of the conversation and extract the final agreed price if so. \\
    
    \textbf{Examples:} \\
    \textit{Conversation:} Buyer: Can you do 630? Seller: I can meet at 635. \\
    \textit{Question:} Deal reached? \textit{Answer:} \{"deal": false, "price": null\} \\
    
    \textit{Conversation:} Buyer: Take 350? Seller: 375. Buyer: 375 works. \\
    \textit{Question:} Deal reached? \textit{Answer:} \{"deal": true, "price": 375\} \\
    
    \textit{Conversation:} Buyer: I can do 1650. Seller: 1650 works. Deal. \\
    \textit{Question:} Deal reached? \textit{Answer:} \{"deal": true, "price": 1650\} \\
    
    \textit{Conversation:} Buyer: 1630? Seller: Reluctantly agree to 1650 final offer. \\
    \textit{Question:} Deal reached? \textit{Answer:} \{"deal": false, "price": null\} \\
    
    Now judge this conversation: \\
    $[$Conversation History$]$ \\
    Answer: \\
    \hline
    \end{tabular}
    \caption{Prompts for critic model in CB task}
    \label{tab:cb_critic_prompt}
\end{table*}

\begin{table*}[h!]
    \centering
    \renewcommand{\arraystretch}{1.2}
    \begin{tabular}{|p{0.95\linewidth}|}
    \hline
    \textbf{Prompts for ICL-AIF Persuader Model in Persuasion (P4G)} \\
    \hline
    \multicolumn{1}{|c|}{\textit{\textbf{Stage 1: Coach Suggestion Generation}}} \\
    \hline
    \textbf{[System]} \\
    Now enter the role-playing mode. In the following conversation, you will play as a coach in a persuasion game. There will be a persuader who is trying to persuade a persuadee for charity donation. Your task is to read the conversation between the persuader and the persuadee, then provide suggestions to the persuader about how to convince the persuadee to make a donation. \\
    \hline
    \textbf{[User]} \\
    Read the following conversation between the persuader and the persuadee, then give three suggestions to the persuader about how to convince the persuadee to make a donation. Each suggestion should be only one short and succinct sentence. The following is the conversation: {[}conversation{]}  \\
    Question: What are your suggestions? \\
    Answer: \\
    \hline
    \hline
    \multicolumn{1}{|c|}{\textit{\textbf{Stage 2: Persuader Response Generation}}} \\
    \hline
    \textbf{[System]} \\
    You are a Persuader trying to persuade the Persuadee to donate to the charity "Save the Children", which helps children in poverty and war zones. Even small donations (\$1–\$2) can make a difference. \\
    \hline
    \textbf{[User]} \\
    Conversation: $[$Conversation History$]$ \\
    Suggestions: $[$Coach Suggestions$]$ \\
    Write ONLY the Persuader's next utterance (one concise sentence). \\
    \hline
    \end{tabular}
    \caption{Prompts for the ICL-AIF framework (Coach \& Persuader) in the P4G task}
    \label{tab:p4g_iclaif_prompts}
\end{table*}

\begin{table*}[h!]
    \centering
    \renewcommand{\arraystretch}{1.2}
    \begin{tabular}{|p{0.95\linewidth}|}
    \hline
    \textbf{Prompts for ProCoT Persuader Model (P4G)} \\
    \hline
    \textbf{[System]} \\
    You are a Persuader trying to persuade the Persuadee to donate to the charity "Save the Children", which helps children in poverty and war zones. Even small donations (\$1–\$2) can make a difference. \\
    \hline
    \textbf{[User]} \\
    Given the conversation history, write ONE response with the exact format below. \\

    Conversation : $[$Conversation History$]$ \\
    
    Strategies (A–J): \\
    A Logical | B Emotion | C Credibility | D Foot-in-the-door | E Self-modeling \\
    F Personal story | G Donation info | H Source inquiry | I Task inquiry | J Personal inquiry \\
    
    Format: \\
    Progress analysis: $($max 2 sentences$)$ \\
    Next-turn goal: $($1 sentence$)$ \\
    Selected persuasion strategies: $($A-J, comma-separated$)$ \\
    Persuader response: $($one concise sentence$)$ \\
    
    Do not output anything other than the required format. \\
    \hline
    \end{tabular}
    \caption{Prompts for the ProCoT Persuader model in the P4G task}
    \label{tab:p4g_procot_prompt}
\end{table*}

\begin{table*}[h!]
    \centering
    \renewcommand{\arraystretch}{1.2}
    \begin{tabular}{|p{0.95\linewidth}|}
    \hline
    \textbf{Prompts for Proactive Persuader Model (P4G)} \\
    \hline
    \textbf{[System]} \\
    You are a Persuader trying to persuade the Persuadee to donate to the charity "Save the Children", which helps children in poverty and war zones. Even small donations (\$1–\$2) can make a difference. \\
    \hline
    \textbf{[User]} \\
    Given the conversation history, select the most appropriate persuasion strategy letter(s) from the list below, then write the Persuader’s next utterance. \\

    Conversation history:$[$Conversation History$]$ \\
    
    Strategies: \\
    A Logical appeal | B Emotion appeal | C Credibility appeal | D Foot-in-the-door \\
    E Self-modeling | F Personal story | G Donation information \\
    H Source-related inquiry | I Task-related inquiry | J Personal-related inquiry \\
    
    Output format (must follow exactly): \\
    Selected persuasion strategies: $($A-J, comma-separated$)$ \\
    Persuader response: $($one concise sentence$)$ \\
    
    Do not output anything other than the required format. \\
    \hline
    \end{tabular}
    \caption{Prompts for the Proactive Persuader model in the P4G task}
    \label{tab:p4g_proactive_prompt}
\end{table*}


\begin{table*}[h!]
    \centering
    \renewcommand{\arraystretch}{1.2}
    \begin{tabular}{|p{0.95\linewidth}|}
    \hline
    \textbf{Prompts for ICL-AIF Buyer Model in Negotiation (CB)} \\
    \hline
    \multicolumn{1}{|c|}{\textit{\textbf{Stage 1: Coach Suggestion Generation}}} \\
    \hline
    \textbf{[System]} \\
    Now enter the role-playing mode. In the following conversation, you will play as a coach in a bargain game. There will be a buyer and a seller bargaining about a product price. \\
    \hline
    \textbf{[User]} \\
    Read the following conversation between the buyer and the seller, then give 3 suggestions to the buyer about how to negotiate more effectively and persuade the seller to accept a better price. Each suggestion should be only one short and succinct sentence. The following is the conversation: \\
    $[$Conversation History$]$ \\
    Question: What are your suggestions? \\
    Answer: \\
    \hline
    \hline
    \multicolumn{1}{|c|}{\textit{\textbf{Stage 2: Buyer Response Generation}}} \\
    \hline
    \textbf{[System]} \\
    You are a Buyer negotiating the price of a $[$Item Name$]$. \\
    Your target price is $[$Target Price$]$. \\
    Rules: \\
    - Reply in 1-2 natural sentences. \\
    - You MUST use the coach's advice to decide your next move. \\
    \hline
    \textbf{[User]} \\
    Item description: $[$Item Description$]$ \\
    Conversation so far: $[$Conversation History$]$ \\
    Coach suggestions: $[$Coach Suggestions$]$ \\
    Write ONLY the buyer's next reply (1-2 sentences). \\
    \hline
    \end{tabular}
    \caption{Prompts for the ICL-AIF framework (Coach \& Buyer) in the CB task}
    \label{tab:cb_iclaif_prompts}
\end{table*}

\begin{table*}[h!]
    \centering
    \renewcommand{\arraystretch}{1.2}
    \begin{tabular}{|p{0.95\linewidth}|}
    \hline
    \textbf{Prompts for ProCoT Buyer Model (CB)} \\
    \hline
    \textbf{[System]} \\
    You are a Buyer negotiating the price of a $[$Item Name$]$. \\
    Your target price is $[$Target Price$]$. \\
    
    Reply in 1–2 short, natural sentences. \\
    \hline
    \textbf{[User]} \\
    Item description: $[$Item Description$]$ \\
    
    Conversation so far: $[$Conversation History$]$ \\

    Write the Buyer's next reply using the EXACT format below. \\
    Buyer negotiation strategies (A–K): \\
    A greet | B ask\_question | C answer\_question | D propose\_first\_price \\
    E propose\_counter\_price | F use\_comparatives | G confirm\_information \\
    H affirm\_confirmation | I deny\_confirmation | J agree\_with\_proposal \\
    K disagree\_with\_proposal \\
    
    Format (must follow exactly): \\
    Progress analysis: $($max 2 sentences$)$ \\
    Next-turn goal: $($1 sentence$)$ \\
    Which one is the most appropriate dialogue strategy: $($A-K, comma-separated$)$ \\
    Buyer response: $($1-2 concise sentence$)$ \\
    
    Do not output anything other than the required format. \\
    \hline
    \end{tabular}
    \caption{Prompts for the ProCoT Buyer model in the CB task}
    \label{tab:cb_procot_buyer_prompt}
\end{table*}


\begin{table*}[h!]
    \centering
    \renewcommand{\arraystretch}{1.2}
    \begin{tabular}{|p{0.95\linewidth}|}
    \hline
    \textbf{Prompts for Proactive Buyer Model (CB)} \\
    \hline
    \textbf{[System]} \\
    You are a Buyer negotiating the price of a $[$Item Name$]$. \\
    Your target price is $[$Target Price$]$. \\
    Reply in 1–2 short, natural sentences. \\
    \hline
    \textbf{[User]} \\
    Item description: $[$Item Description$]$ \\
    Conversation so far: $[$Conversation History$]$ \\
    Buyer negotiation strategies: \\
    A greet | B ask\_question | C answer\_question | D propose\_first\_price \\
    E propose\_counter\_price | F use\_comparatives | G confirm\_information \\
    H affirm\_confirmation | I deny\_confirmation | J agree\_with\_proposal \\
    K disagree\_with\_proposal \\
    
    Output format (must follow exactly): \\
    Which one is the most appropriate dialogue strategy: $($A-K, comma-separated$)$ \\
    Buyer response: $($1-2 concise sentence$)$ \\
    Do not output anything else. \\
    \hline
    \end{tabular}
    \caption{Prompts for the Proactive Buyer model in the CB task}
    \label{tab:cb_proactive_buyer_prompt}
\end{table*}

\begin{table*}[h]
    \centering
    \renewcommand{\arraystretch}{1.2}
    \begin{tabular}{|p{0.95\linewidth}|}
    \hline
    \textbf{Inference-Time Strategy Generation Prompts} \\
    \hline
    \textbf{[Short-Horizon Strategy Prompt]} \\
    \textbf{System:} You are an expert persuasion strategist. \\
    Given several retrieved micro-principles and the user's latest message, provide ONE high-level strategy
    (1--2 sentences) describing what the persuader should do next. \\
    Constraints: focus on the next action only; do not restate conditions (e.g., ``when''); do not copy or paraphrase the principles. \\
    \\
    \textbf{User:} \\
    User's latest message: $[$Latest User Utterance$]$ \\
    Retrieved micro-principles (Top-5): $[$Principles$]$ \\
    Output: ONE concise next-step strategy (1--2 sentences). \\
    \hline
    \textbf{[Long-Horizon Strategy Prompt]} \\
    \textbf{System:} You are an expert persuasion strategist specialized in long-term planning. \\
    Given the recent dialogue context and a strategy chain extracted from successful dialogues,
    provide a high-level summary (1--2 sentences) describing how the persuader should plan their strategy over time. \\
    Constraints: focus on strategic direction and psychological progression; do not list operational steps; do not restate the chain verbatim. \\
    \\
    \textbf{User:} \\
    Recent dialogue context: $[$Recent Dialogue History$]$ \\
    Retrieved strategy chain (Top-1): $[$Strategy Chain$]$ \\
    Output: ONE concise long-term strategy summary (1--2 sentences). \\
    \hline
    \end{tabular}
    \caption{Prompts used for inference-time short-horizon and long-horizon strategy generation}
    \label{tab:inference_strategy_prompts}
\end{table*}


\begin{table*}[h]
    \centering
    \renewcommand{\arraystretch}{1.2}
    \begin{tabular}{|p{0.95\linewidth}|}
    \hline
    \textbf{Prompt for persona generation} \\
    \hline
    \textbf{Instruction.} \\
    You are given a set of persona attributes. Your task is to integrate these
    attributes and rewrite them into a cohesive, natural persona description
    written in the second person. \\
    
    Persona Attributes. \\
    Big-Five Personality: $[$big five$]$ \\
    Decision-Making Style: $[$decision style$]$ \\
    
    \textbf{Requirements.}
    \begin{itemize}[leftmargin=1.5em, itemsep=1pt]
        \item Do not explicitly mention or repeat the personality labels.
        \item Express the persona implicitly through behaviors, preferences,
        emotional tendencies, communication style, and decision-making patterns.
        \item The description should be fluent, natural, and easy to understand.
        \item The length should be approximately 80--120 words.
    \end{itemize}
    
    \textbf{Style Reference} \\
    \emph{``You are a 28-year-old software developer who enjoys exploring new ideas
    and unfamiliar experiences. In daily life, you tend to weigh different options
    carefully before committing to a decision. When others speak with you, you
    respond thoughtfully and value clear, logical reasoning. Although you take time
    to reflect, you can be persuaded by ideas that feel both meaningful and
    well-justified.''} \\
    
    \textbf{Output.} \\
    Generate a concise persona description following the above instructions. \\
    
    \hline
    \end{tabular}
    \caption{Prompt used to construct discrete personas}
    \label{tab:ICL-AIF_p4g_persuader_prompt}
\end{table*}

\end{document}